\documentclass[10pt,twocolumn,letterpaper]{article}

\usepackage[pagenumbers]{cvpr}

\usepackage[accsupp]{axessibility}
\usepackage{graphicx}
\usepackage{amsmath}
\usepackage{amssymb}
\usepackage{booktabs}
\usepackage{bm}
\usepackage{multirow}

\usepackage[table,xcdraw]{xcolor}
\usepackage{amssymb}
\usepackage{pifont}
\usepackage{bbm}

\newcommand\customparagraph[1]{\vspace{0.5em}\noindent\textbf{#1}}
\renewcommand{\mathbbm}[1]{\text{\usefont{U}{bbm}{m}{n}#1}}

\newcommand{\wincat}[1]{\textcolor{black}{\textbf{#1}}}

\newcommand{\cmark}{\ding{51}}
\newcommand{\xmark}{\ding{55}}

\usepackage{xspace}

\definecolor{lightgray}{rgb}{0.835, 0.835, 0.835}
\definecolor{lightergray}{rgb}{0.935, 0.935, 0.935}

\definecolor{lighterblue}{rgb}{0.93, 0.97, 1.0}
\definecolor{lightblue}{rgb}{0.83, 0.90, 1.0}

\newcolumntype{a}{>{\columncolor{lightgray}}c}

\usepackage[pagebackref,breaklinks,colorlinks]{hyperref}

\usepackage[capitalize]{cleveref}
\crefname{section}{Sec.}{Secs.}
\Crefname{section}{Section}{Sections}
\Crefname{table}{Table}{Tables}
\crefname{table}{Tab.}{Tabs.}

\begin{document}

\title{Neural Voting Field for Camera-Space 3D Hand Pose Estimation}

\newcommand{\namesep}{\hspace{4.0em}}
\newcommand{\institutionsep}{\hspace{4.0em}}
\author{
    Lin Huang{\small $^{1*}$}\namesep
    Chung-Ching Lin{\small $^{2}$}\namesep
    Kevin Lin{\small $^{2}$}\namesep
    Lin Liang{\small $^{2}$}\\
    Lijuan Wang{\small $^{2}$}\namesep
    Junsong Yuan{\small $^{1}$}\namesep
    Zicheng Liu{\small $^{2}$}\\ 
    {\small $^{1}$}University at Buffalo\institutionsep
    {\small $^{2}$}Microsoft\\
    \texttt{\href{https://linhuang17.github.io/NVF}{\small https://linhuang17.github.io/NVF}}
}

\twocolumn[{
\renewcommand\twocolumn[1][]{#1}
\maketitle
\begin{center}  
    \vspace{-2.5ex}
    \centering
    \captionsetup{type=figure}
         \hspace{0.0cm} {\footnotesize RGB input} \hspace{0.3cm} {\footnotesize 3D hand mesh}  \hspace{0.7cm}  {\footnotesize Wrist} \hspace{0.7cm}   {\footnotesize Thumb root} \hspace{0.5cm}  {\footnotesize Index tip} \hspace{0.7cm}  {\footnotesize Middle root} \hspace{0.6cm}  {\footnotesize Ring tip} \hspace{0.8cm}  {\footnotesize Pinky root}  \hspace{0.35cm} {\footnotesize 3D hand pose}
    \vspace{-1ex}
    \includegraphics[width=1.0\linewidth]{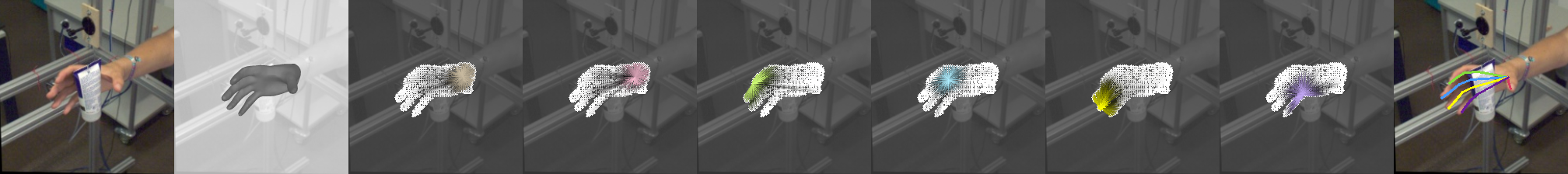}
    \vspace{-1.5ex}
        \hspace{-0.1cm} {\footnotesize (a)} \hspace{1.45cm}   {\footnotesize (b)}  \hspace{1.5cm}  {\footnotesize (c)} \hspace{1.4cm}  {\footnotesize (d)} \hspace{1.4cm}  {\footnotesize (e)} \hspace{1.5cm}  {\footnotesize (f)} \hspace{1.5cm}  {\footnotesize (g)} \hspace{1.3cm}  {\footnotesize (h)} \hspace{1.5cm}  {\footnotesize (i)}
    \caption{We present \textbf{Neural Voting Field (NVF)}, a 3D implicit function for camera-space 3D hand pose estimation from an RGB image.
    Given an RGB input (a), for a 3D point in camera space, NVF can predict its signed distance to hand surface and a set of 4D offset vectors (1D voting weight and 3D directional vector to each joint). 3D pose is then obtained via dense 3D point-wise voting. 
    Rendered predictions:
    (b) Hand mesh generated by Marching Cubes from the predicted signed distances 
    (c-h) Predicted 1D voting weight from each predicted near-surface 3D point to a joint. (Brighter line means larger weight, darker or no line means smaller weight)
    (i) Estimated 3D hand pose.}
    \label{fig:teaser}
\end{center}
}]

{\let\thefootnote\relax\footnotetext{$^{*}$Work done during Lin Huang's internship with Microsoft.}}
\begin{abstract}
We present a unified framework for camera-space 3D hand pose estimation from a single RGB image based on 3D implicit representation. As opposed to recent works, most of which first adopt holistic or pixel-level dense regression to obtain relative 3D hand pose and then follow with complex second-stage operations for 3D global root or scale recovery, we propose a novel unified 3D dense regression scheme to estimate camera-space 3D hand pose via dense 3D point-wise voting in camera frustum. Through direct dense modeling in 3D domain inspired by Pixel-aligned Implicit Functions for 3D detailed reconstruction, our proposed Neural Voting Field (NVF) fully models 3D dense local evidence and hand global geometry, helping to alleviate common 2D-to-3D ambiguities. Specifically, for a 3D query point in camera frustum and its pixel-aligned image feature, NVF, represented by a Multi-Layer Perceptron, regresses: (i) its signed distance to the hand surface; (ii) a set of 4D offset vectors (1D voting weight and 3D directional vector to each hand joint). Following a vote-casting scheme, 4D offset vectors from near-surface points are selected to calculate the 3D hand joint coordinates by a weighted average. Experiments demonstrate that NVF outperforms existing state-of-the-art algorithms on FreiHAND dataset for camera-space 3D hand pose estimation. We also adapt NVF to the classic task of root-relative 3D hand pose estimation, for which NVF also obtains state-of-the-art results on HO3D dataset.
\end{abstract}
\vspace{-5ex}
\section{Introduction}
\label{sec:intro}
Monocular 3D hand pose estimation, which aims to recover 3D locations of hand joints from an RGB image, has attracted enormous attention and made remarkable progress in recent years. 
As a long-standing task in computer vision, it remains challenging due to its highly articulated structure, large variations in orientations, severe (self-)occlusion, and inherent 2D-to-3D scale and depth ambiguity.

Owing to the aforementioned difficulties, most existing works~\cite{zimmermann2017learning,spurr2018cross,ge20193d,boukhayma20193d,zhang2019end,cai2019exploiting,yang2019aligning,kulon2020weakly,lin2021end,chen2021i2uv,lin2021mesh,chen2022mobrecon,tse2022collaborative,hampali2022keypoint,park2022handoccnet,huang2021survey} focused on one aspect of this general problem, which is to estimate root-relative 3D hand pose (\ie, 3D joint coordinates relative to a pre-defined root joint, such as hand wrist). While accurate 2D-to-3D root-relative pose estimation is essential for numerous applications in Virtual/Augmented Reality, there are various interactive tasks in which having root-relative hand joint coordinates alone is insufficient. For instance, being able to recover camera-space 3D hand joint coordinates in an AR view enables the user to directly use hands to manipulate virtual objects moving in 3D space.

To recover robust camera-space 3D hand pose, there are two key design elements: (1) the ability to exploit dense local evidence. Specifically, as demonstrated in previous works~\cite{wei2016convolutional,newell2016stacked,iqbal2018hand,wan2018dense,xiong2019a2j,fang2020jgr,ge2018point,li2019point,huang2020hand,pavlakos2017coarse,moon2020i2l,saito2019pifu}, dense regression-based methods are more effective than holistic regression-based counterparts for handling highly articulated 3D pose structure, attributed to its ability to maintain the input data spatial structure and fully exploit local evidence; (2) the ability to reason 3D hand global geometry.
As shown in previous literature~\cite{dijk2019neural,iqbal2018hand,lepetit2009epnp}, given 2D evidence and camera intrinsic parameters, reasonable understanding towards target object 3D structure/geometry is crucial to alleviate 2D-to-3D depth ambiguity, which is the key to accurately locate 3D hand pose in camera space.

To fully integrate both elements into our algorithm design in a unified manner, we connect with Pixel-aligned Implicit Function (PIFu)~\cite{saito2019pifu,saito2020pifuhd,huang2020arch,zheng2021pamir,huang2022neural}.
Through direct dense modeling in 3D domain with pixel-aligned local features, PIFu-based methods reconstruct highly detailed 3D human geometry from an RGB image in a unified way, showing its ability to model high-frequency local details such as clothing wrinkles while generating complete global geometry including largely occluded region such as the back of a person. 
Inspired by these results, we propose a novel unified 3D dense regression scheme based on a 3D implicit function for robust camera-space 3D hand pose estimation.
Specifically, for each of the 3D query points densely sampled in camera frustum and its pixel-aligned image feature, unlike PIFu predicting occupancy value for each point, our proposed Neural Voting Field (NVF) regresses: (i) the signed distance between the point and the hand surface; (ii) a set of 4D offset vectors (1D voting weight and 3D directional vector from the point to each joint).
Following a vote-casting scheme, 4D offset vectors from near-surface points (\ie, points for which the predicted signed distance is below a threshold) are selected to calculate the 3D hand joint coordinates by a weighted average.

\begin{table}[t]
\footnotesize
\centering	
\begin{tabular}{ccc}
\hline
{Method} & {First Stage} & {Second Stage} \\ \hline
\rowcolor{lightblue}
Iqbal \etal~\cite{iqbal2018hand}               & 2D-Dense                      & Scale Estimation   \\ \hline
ObMan~\cite{hasson19_obman}                    & Holistic                      & Root Estimation    \\ \hline
\rowcolor{lightblue}
I2L-MeshNet~\cite{moon2020i2l}                 & 1D-Dense                      & Root Depth Estimation \\ \hline
CMR~\cite{chen2021camera}                      & 2D-Dense+SpiralConv           & Registration      \\ \hline
\rowcolor{lightblue}
Hasson \etal~\cite{hasson2021towards}          & Holistic                      & Model Fitting      \\ \hline
NVF (Ours)                                     & Unified 3D-Dense              & Weighted Average        \\ \hline
\end{tabular}
\vspace{-1.5ex}
\caption{\label{table:survey}\textbf{Comparison of representative absolute 3D hand pose estimation schemes.} 
Please refer to Sec.~\ref{sec:related_cs} for more details.
}
\vspace{-4ex}
\end{table}
Most existing works for camera-space 3D hand pose estimation, as shown in Tab.~\ref{table:survey}, follow a two-stage estimation scheme. They first adopt holistic or pixel-level dense regression to obtain 2D and relative 3D hand poses and then follow with complex second-stage processing such as fitting, registration, using a separate network for 3D global root location or scale estimation. 
NVF instead provides a unified solution via direct dense modeling in 3D camera space followed by a simple weighted average operation, which enables reasoning about 3D dense local evidence and hand global geometry. As shown in Fig.~\ref{fig:teaser}, NVF makes solid 3D point-wise prediction and overall distribution of signed distance and voting weight even in highly occluded regions, leading to accurate camera-space pose estimation.
In Sec.~\ref{sec:exp}, we show that NVF noticeably outperforms two baselines based on holistic regression and 2D dense regression. Besides, NVF exhibits state-of-the-art performance for the task of camera-space 3D hand pose estimation on FreiHAND dataset. We also adapt NVF to the classic task of root-relative 3D hand pose estimation, for which NVF also achieves state-of-the-art results on HO3D dataset.

Since estimating absolute 3D pose from an RGB image is an ill-posed problem due to scale and depth ambiguity ~\cite{dijk2019neural,iqbal2018hand}, in Sec.~\ref{sec:base_study}, we also provide ablation analysis on hand scale based on results from NVF and the baselines.

This work makes the following contributions:
\vspace{-1.0ex}
\begin{enumerate}
\item We propose Neural Voting Field (NVF), as the first 3D implicit representation-based unified solution to estimate camera-space 3D hand pose. 
\vspace{-1.0ex}
\item NVF follows a novel unified 3D dense regression scheme to estimate camera-space 3D hand pose via dense 3D point-wise voting in camera frustum. 
\vspace{-1.0ex}
\item NVF outperforms baseline methods based on holistic and 2D dense regression and achieves state-of-the-art results on absolute and relative hand pose estimation.
\end{enumerate}

\begin{figure*}[!t]
    \centering
	\includegraphics[width=0.95\linewidth]{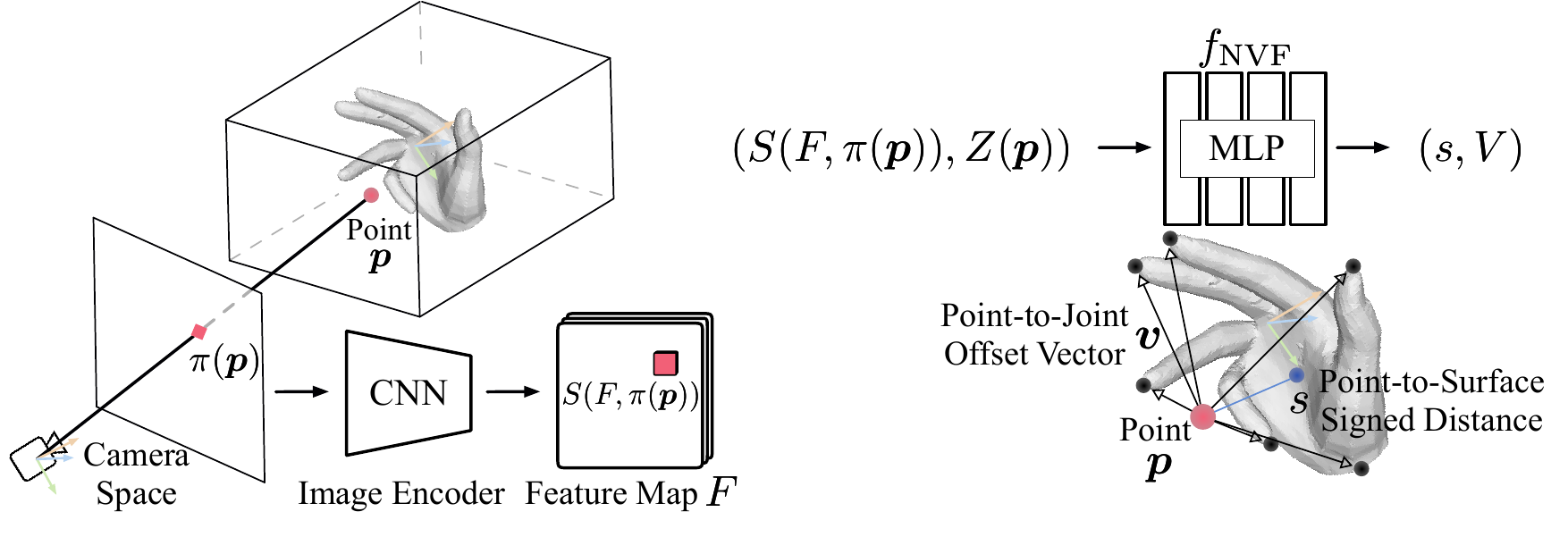}
	\vspace{-2.5ex}
	\caption{Overview of \textbf{Neural Voting Field (NVF)}: We first extract a feature map $F$ from the input RGB image $I$ by an image encoder. Then, for a 3D query point $\boldsymbol{p}$ sampled in camera space, we define an implicit function $f_\text{NVF}$ realized by a MLP, which maps from pixel-aligned image feature $S(F,\pi(\boldsymbol{p}))$ and depth $Z(\boldsymbol{p})$ of $\boldsymbol{p}$ to its signed distance $s$ and a set of 4D offset vectors $V$. Each 4D offset vector $\boldsymbol{v}$ contains a 1D voting weight and a 3D directional vector to each joint. The 3D hand pose is then obtained by dense 3D point-wise voting.}
	\label{fig:nvf}
	\vspace{-3ex}
\end{figure*}

\vspace{-2ex}
\section{Related Work}
\label{sec:related}
\subsection{Camera-Space 3D Hand Pose Estimation}
\label{sec:related_cs}
As shown in Tab.~\ref{table:survey}, most existing works for monocular RGB input-based camera-space 3D hand pose estimation generally follow a two-stage estimation scheme.
Iqbal \etal~\cite{iqbal2018hand} recovers 3D hand pose in camera space up to a scaling factor by solving a quadratic equation given the camera intrinsic parameters and the estimated scale-normalized and root-relative 2.5D pose via 2D Dense regression. Extra global scale estimation is required for camera-space absolute 3D hand pose recovery.
Another line of work~\cite{hasson2021towards,hasson19_obman,chen2021camera} first obtains 2D pose or root-relative 3D hand pose estimations via 2D dense or holistic regression. Then, Hasson \etal~\cite{hasson2021towards} fits a 3D hand model to the initial estimations and recovers 3D hand pose in camera space from the input RGB video.
Both ObMan~\cite{hasson19_obman} and CMR~\cite{chen2021camera} work on a single RGB input and directly recover 3D global root from the initial 2D pose and root-relative 3D pose/mesh estimations.
I2L-MeshNet~\cite{moon2020i2l} proposes a 1D dense regression scheme to predict per-lixel likelihood on 1D heat map for recovering root-relative 2.5D pose, and then uses a separate network (\ie, RootNet~\cite{moon2019camera}) for the absolute root depth estimation.

\subsection{Root-relative 3D Hand Pose Estimation}
\label{sec:related_rs}
For single RGB input-based root-relative 3D hand pose estimation, some early methods~\cite{zimmermann2017learning,spurr2018cross,yang2019aligning} adopt classic frameworks such as Convolutional Neural Network and Multi-Layer Perceptron in combination with holistic or dense regression scheme for scale-invariant and root-relative 3D hand joint positions recovery.
After the introduction of the 3D parametric MANO~\cite{romero2022embodied} model, tremendous literature~\cite{hasson19_obman,boukhayma20193d,baek2019pushing,zhang2019end,chen2021model,park2022handoccnet,tse2022collaborative} attempts to integrate MANO into the learning pipeline and directly regress MANO parameters from the image.
Meanwhile, MANO model-free approaches~\cite{chen2021i2uv} continue to develop and more works also start to investigate the dependencies among hand joints as hands are inherently structured. Thus, geometric deep learning-based models (\eg, Graph Convolution, Spiral Convolution)~\cite{ge20193d,cai2019exploiting,kulon2020weakly,chen2021camera,chen2022mobrecon,tse2022collaborative} and Transformer-based models~\cite{lin2021end,lin2021mesh,hampali2022keypoint,park2022handoccnet} have also been proposed.

\subsection{3D Implicit Representation}
\label{sec:related_implicit}
Recent methods~\cite{occnet:mescheder:cvpr19,lif:chen:cvpr19,liu2021fully,deepsdf2019park,igr:gropp:icml20,sal:atzmon:cvpr20,sald:atzmon:iclr21,metasdf:sitzmann:2019} have shown that, as a continuous and differentiable function realized by Multi-Layer Perceptron, memory efficient deep implicit function can represent detailed 3D shapes with arbitrary topology and have no resolution limitations.
It can also be extended to RGB image-based 3D reconstruction~\cite{texfield:oechsle:iccv19,occnet:mescheder:cvpr19,deepsdf2019park,disn:xu:neurips19}.
Grasping Field~\cite{graspfield:3dv2020} proposes to reconstruct hands grasping object via predicting signed distances for hand and object given the input point cloud or RGB image. 
Unlike previous implicit functions relying on a global feature vector, given an input RGB image with cleanly segmented foreground, Pixel-aligned Implicit Function~\cite{saito2019pifu} (PIFu) utilizes pixel-aligned image feature and depth value of each 3D point for occupancy prediction. 
PIFu-based methods have shown the ability to generate high-fidelity 3D human geometry with more clothes details~\cite{saito2020pifuhd,huang2020arch,zheng2021pamir,kulkarni2021s} and robust 6D pose estimation.
For instance, Neural Correspondence Field~\cite{huang2022neural} predicts signed distance and 3D object coordinate for each 3D point to establish dense 3D-3D correspondences for robust 6D object pose estimation.
\section{The Proposed Method}
\label{sec:method}
\subsection{Preliminaries}
\vspace{-1ex}
\customparagraph{Pixel-Aligned Implicit Function (PIFu)~\cite{saito2019pifu}.}
PIFu, by learning an implicit function in 3D domain with pixel-aligned features, generates detailed 3D human geometry including largely occluded regions from a segmented RGB image. Specifically, given a segmented image $I:\mathbb{R}^2\mapsto\mathbb{R}^3$, PIFu first extracts a feature map $F:\mathbb{R}^2\mapsto\mathbb{R}^{C}$ with $C$ channels by a fully convolutional image encoder. Then, for a 3D point $\boldsymbol{p}\in\mathbb{R}^3$ in the camera space and its 2D projection $\pi(\boldsymbol{p}):\mathbb{R}^3\mapsto\mathbb{R}^2$ on $I$, the implicit function, represented by a Multi-Layer Perceptron (MLP) $f_\text{PIFu}: \mathbb{R}^{C} \times \mathbb{R}\mapsto\mathbb{R} \text{~as~}  f_\text{PIFu}(S(F,\pi(\boldsymbol{p})), Z(\boldsymbol{p})) = y$ maps the pixel-aligned feature $S(F,\pi(\boldsymbol{p}))$ and the depth $Z(\boldsymbol{p})$ of $\boldsymbol{p}$ to the occupancy $y\in[0,1]$ ($1$ means $\boldsymbol{p}$ is inside the human surface and $0$ means otherwise). $S$ is the sampling function (\eg, bilinear interpolation) to sample the value of $F$ at pixel $\pi(\boldsymbol{p})$. 

\customparagraph{Signed Distance Function (SDF)~\cite{tsdf:curless96, deepsdf2019park}.} Given a 3D shape $M\in \mathcal{M}$, a signed distance function is a continuous function $\phi: \mathbb{R}^{3}\times\mathcal{M}\mapsto\mathbb{R} \text{~as~} \phi(\boldsymbol{p},M)=s$, mapping a 3D point $\boldsymbol{p}$ to the signed distance $s$ between $\boldsymbol{p}$ and the shape surface. 
The sign denotes if $\boldsymbol{p}$ is outside (positive) or inside (negative) of a watertight surface, with the iso-surface $\phi(\boldsymbol{p},M)=0$ implicitly representing the surface.

\vspace{-1ex}
\subsection{Dense Offset-based Pose Re-Parameterization}
As opposed to holistic regression directly mapping the input data to sparse and absolute joint positions and 2D dense regression which mainly exploits the 2D input local evidence, our method aims to directly model in 3D domain and thus fully exploit the 3D dense local evidence and hand global structure.
To achieve this goal, given a common sparse and explicit pose representation (\ie, a set of 3D coordinates of $T$ hand joints) as $J=\{\boldsymbol{j}_t\}_{t=1}^{T}\in\mathcal{J}$, $\boldsymbol{j}_t\in\mathbb{R}^{3}$, we parameterize it as a set of dense 3D point-wise offset-based representation.
As shown in previous works~\cite{wan2018dense,ge2018point}, it is non-ideal to directly learn the 3D offset vector from each 3D point to a joint due to the large variance of offsets and far away joints are also beyond the scope of receptive field. 
Thus, given a camera-space 3D hand pose $J$ and shape $M$, we define the pose parameterization as a function $\psi$, mapping from a 3D point $\boldsymbol{p}$ in the camera space to a set of 4D offset vectors
$V=\{\boldsymbol{v}_t\}_{t=1}^{T}$, $\boldsymbol{v}_t\in\mathbb{R}^{4}$:
\begin{equation}
    \psi: \mathbb{R}^{3} \times \mathcal{J} \times \mathcal{M}  \mapsto \mathbb{R}^{T \times 4}     \text{~~as~~}
    \psi(\boldsymbol{p},J,M)=V.
\label{eq:pose_repre}
\end{equation}
Each offset vector $\boldsymbol{v}_t=(w_t,\boldsymbol{d}_t)$ consists of a 1D voting weight $w_t\in\mathbb{R}$ as point-wise closeness to hand joint $j_t$:
\begin{equation}
w_t=
\left\{ {\begin{array}{*{20}{ll}}
    1-\frac{\|\boldsymbol{j}_t-\boldsymbol{p}\|_2}{r} \ \ 
    |s|<\delta~\text{and}~\|\boldsymbol{j}_t-\boldsymbol{p}\|_2 \le r~\text{and}~\boldsymbol{p}\in{B}_t^K,\\
     \ \ \ \  \ \ \ \ \ \ 0 \ \ \ \ \ \ \ \ \ \ \ \text{otherwise;}
	\end{array}} \right.
\label{eq:vote_weight}
\end{equation}
and a 3D unit directional vector $\boldsymbol{d}_t\in\mathbb{R}^{3}$ as point-wise direction to hand joint $j_t$: 
\begin{equation}
\boldsymbol{d}_t=
\left\{ {\begin{array}{*{20}{ll}}
    \frac{\boldsymbol{j}_t-\boldsymbol{p}}{\|\boldsymbol{j}_t-\boldsymbol{p}\|_2} \ \  |s|<\delta~\text{and}~\|\boldsymbol{j}_t-\boldsymbol{p}\|_2 \le r~\text{and}~\boldsymbol{p}\in{B}_t^K,\\
     \ \ \ \ \ \ \bm{0} \ \ \ \ \ \ \ \ \text{otherwise;}
	\end{array}} \right.
\label{eq:dir_vec}
\end{equation}
where $s=\phi(\boldsymbol{p},M)$ and $|s|<\delta$ means points that are within clamping distance $\delta$ from the hand surface.
$r$ denotes the radius of a 3D ball centered at each joint and $B_t^K$ denotes the set of $K$-Nearest Neighbors (${K}$NN) to joint $\boldsymbol{j}_t$ among all the near-surface points (\ie,  points with $|s|<\delta$).
Thus, among all the 3D points that are in the hand surface vicinity, we assign the non-zero 4D offset vector $\boldsymbol{v}_t$ to the $K$ nearest points inside the 3D ball with radius $r$ of joint $\boldsymbol{j}_t$.

\subsection{Neural Voting Field}
\label{sec:nvf}
To fully exploit the 3D local evidence and hand global geometry for camera-space 3D hand pose estimation, based on the aforementioned signed distance function and pose parameterization, we unify both modalities using a pixel-aligned implicit function. Following the proposed pipeline in Fig.~\ref{fig:nvf}, we first extract a $C$-channel feature map $F$ from the input RGB image $I$ by a hourglass network $g(I;\boldsymbol{\eta})$ as used in~\cite{saito2019pifu,huang2022neural}. Then, for a 3D query point $\boldsymbol{p}$ sampled in camera space, we define a continuous implicit function $f_\text{NVF}$ realized by a MLP, which maps from pixel-aligned image feature $S(F,\pi(\boldsymbol{p}))$ and depth $Z(\boldsymbol{p})$ of point $\boldsymbol{p}$ to its signed distance  $s$ and a set of 4D offset vectors $V$:
\begin{equation}
{\begin{array}{*{20}{ll}}
    f_\text{NVF}: \mathbb{R}^C \times \mathbb{R} \mapsto \mathbb{R} \times \mathbb{R}^{T\times4} \text{~~~as~~~} \\
    f_\text{NVF}\big(S(F,\pi(\boldsymbol{p})), Z(\boldsymbol{p}); \boldsymbol{\theta}\big) = \big(s,V\big),
	\end{array}} 
\label{eq:NVF}
\end{equation}
where $\boldsymbol{\eta}$ and $\boldsymbol{\theta}$ are the parameters of the hourglass $g$ and MLP $f_\text{NVF}$, respectively. We use a pinhole camera model for 3D-to-2D projection. For each 3D point, besides the set of 4D offset vectors, NVF also predicts its signed distance to the hand surface, helping to learn the global hand geometry and also to select near-surface 3D points by thresholding the signed distances. The near-surface 3D points are then considered as valid 3D voters to be used in the voting stage.

\customparagraph{Optimization.}
During training, we sample $N$ 3D points $P=\{\boldsymbol{p}_n\}_{n=1}^{N}$ in camera frustum for each image and apply direct supervision to both predicted signed distance $s_n$ as $L_s$ and the predicted set of 4D offset vectors $V_n$ as $L_V$. Given the ground-truth signed distance $\hat{s}_n=\phi(\boldsymbol{p}_n,\hat{M})$ and set of offset vectors $\hat{V}_n=\psi(\boldsymbol{p}_n,\hat{J},\hat{M})$, where $\hat{J}$ and $\hat{M}$ are the ground-truth pose and mesh, 
$L_s$ and $L_V$ are defined as:
\vspace{-3ex}
\begin{align}
    L_s &= \frac{1}{N} \sum_{n=1}^{N}
    \big| \text{clamp}\left(\hat{s}_n, \delta\right) - \text{clamp}\left(s_n, \delta\right) \big|, \\
    L_V &= \frac{1}{N} \sum_{n=1}^{N}
    \mathbbm{1}\left(|\hat{s}_n| < \delta\right)
    H\left(\hat{V}_n,V_n\right),
    \label{eq:loss}
\end{align}
where $\text{clamp}(s,\delta):=\text{min}(\delta,\text{max}(-\delta,s))$ uses $\delta$ to control the distance from the hand surface over which we want to main a metric SDF, following~\cite{deepsdf2019park}. $H$ is the Huber loss~\cite{huber1992robust} and $\mathbbm{1}(\cdot)$ is an indicator function that selects near-surface points (\ie, points that are within clamping distance $\delta$ from the hand surface).
Then, We jointly train the hourglass network $g$ and MLP $f_\text{NVF}$ end-to-end by solving the following optimization problem with $\lambda$ as the balancing weight:
\vspace{-1ex}
\begin{align}\label{eq:all_loss}
    \boldsymbol{\eta}^\star, \boldsymbol{\theta}^\star = \underset{\boldsymbol{\eta,\theta}}{\arg\min} \;
    L_{s} + \lambda L_{V}.
\end{align}
\vspace{-2ex}

\customparagraph{Sampling of 3D Query Points.}
During training, unlike voxel-based methods which require discretization of ground-truth 3D meshes, we directly sample 3D points on the fly from the ground-truth mesh following~\cite{saito2019pifu,huang2022neural}.
To make NVF fully exploit meaningful local evidence and understand the hand global geometry, it is crucial to focus on the hand by sampling the query points more densely around the hand surface during training. 
Thus, given the 3D camera-space hand model, we first sample 12500 points nearby the surface, 1000 points inside the bounding sphere of the model, and 1000 points inside the camera frustum.
Among all these points, we then sample 2500 points inside the model and 2500 points outside to form a set of 5000 points in total for training.
During inference, the points are directly sampled at centers of voxels that fill up the same camera frustum without any ground-truth information used.

\subsection{Dense 3D Point-to-Joint Voting.}
During inference, assuming $N$ 3D points $P=\{\boldsymbol{p}_n\}_{n=1}^{N}$ sampled in camera frustum for each image, for each point $\boldsymbol{p}_n$ that is considered as a valid 3D voter (\ie, the predicted signed distance $s_n$ is below the threshold $\delta$), based on its predicted set of 4D offset vectors $V_n=\{\boldsymbol{v}_t^n\}_{t=1}^{T}$,
we first convert each 4D offset vector $\boldsymbol{v}_t^n=(w_t^n, \boldsymbol{d}_t^n)$ 
into the actual 3D offset $\boldsymbol{o}_t^n$ from $\boldsymbol{p}_n$ to joint $\boldsymbol{j}_t$ based on Eq.~\ref{eq:vote_weight} and Eq.~\ref{eq:dir_vec}:
\vspace{-3ex}
\begin{align}\label{eq:offset}
    \boldsymbol{o}_t^n = \mathbbm{1}\left(|s_n| < \delta\right)[r(1-w_t^n)\boldsymbol{d}_t^n].
\end{align}
Then, each camera-space joint location $\boldsymbol{j}_t$ is calculated simply by a weighted average over each voter's 3D offset prediction: 
\begin{equation}
\boldsymbol{j}_t=
{\sum_{n=1}^N }
\frac{\mathbbm{1}\left(|s_n| < \delta\right){w_t^n}(\boldsymbol{o}_t^n+\boldsymbol{p}_n)}{\sum\nolimits_{n=1}^N {\mathbbm{1}\left(|s_n| < \delta\right){w_t^n}} }.
\label{eq:pose}
\end{equation}
Finally, we obtain the final camera-space 3D hand pose as $J=\{\boldsymbol{j}_t\}_{t=1}^{T}$.
Furthermore, instead of using all the near-surface points as valid voters, we can also only focus on using near-surface points with high predicted voting weight, as a higher voting weight means the point is more confident and closer to a hand joint. 
Therefore, we use voting fraction as a parameter to control the fraction of near-surface points with the highest voting weight for voting. Related ablation studies will be given in Sec.~\ref{sec:ablation}.

\section{Experiments}
\label{sec:exp}
\subsection{Baseline Methods}
\label{sec:baseline}
To demonstrate the effectiveness of our proposed 3D dense regression scheme against classic holistic regression and 2D dense regression, we design two directly comparable baselines that share the same architecture of the hourglass network and the MLP as NVF described in Sec.~\ref{sec:method}. 
Specifically, similar to NVF, both baselines first extract $C$-channel feature map from the input RGB image $I$. Then, unlike our $f_\text{NVF}$ which takes a sampled 3D query point's pixel-aligned feature $S(F,\pi(\boldsymbol{p}))$ and its depth $Z(\boldsymbol{p})$ to predict its signed distance $s$ and the set of 4D offset vectors $V$, the two baselines take the following steps.

\customparagraph{Baseline-Holistic.}
We directly apply a global average pooling $\alpha$ to its feature map $F^{\prime}$ extracted by the hourglass network $g(I;\boldsymbol{\eta^{\prime}})$ and generate a $C$-dimension feature vector $\alpha(F^{\prime})$. The feature vector is then passed through the MLP $f_\text{HOL}: \mathbb{R}^{C}\mapsto\mathbb{R}^{T\times3} \text{~as~}  f_\text{HOL}(\alpha(F^{\prime}); \boldsymbol{\theta^{\prime}}) = J^{\prime}$. $f_\text{HOL}$ directly outputs $T$ 3D hand joint coordinates $J^{\prime}=\{\boldsymbol{j}_t^{\prime}\}_{t=1}^{T}$ in camera space. 
Given the ground-truth pose $\hat{J}$, it is trained by solving the following optimization problem with a loss function $L_{J^{\prime}}=H(\hat{J},J^{\prime})$:
\vspace{-1ex}
\begin{align}\label{eq:bh_loss}
    &\boldsymbol{\eta^{\prime\star}}, \boldsymbol{\theta^{\prime\star}} = \underset{\boldsymbol{\eta^{\prime},\theta^{\prime}}}{\arg\min} \;
    L_{J^{\prime}}.
\end{align}
\vspace{-3ex}

\begin{table*}[t]
\begin{minipage}{\columnwidth}
\centering
\captionsetup{width=1.0\textwidth}
\begin{tabular}{cccccc}
\hline
\multirow{2}{*}{\vspace{2.5ex}Method} & {\begin{tabular}[c]{@{}c@{}}Extra \\ Data\end{tabular}} & {\begin{tabular}[c]{@{}c@{}}Hand \\ Crop\end{tabular}} & {\begin{tabular}[c]{@{}c@{}}Hand \\ Scale\end{tabular}} & {CS-MJE}$\downarrow$ \\ 
\hline
ObMan~\cite{hasson19_obman}                        & -       & \cmark   & \xmark   & 85.2             \\
MANO CNN~\cite{zimmermann2019freihand}             & -       & \cmark   & \xmark   & 71.3             \\
I2L-MeshNet~\cite{moon2020i2l}                     & -       & \cmark   & \xmark   & 60.3            \\
CMR-SG-RN18~\cite{chen2021camera}                  & -       & \cmark   & \xmark   & 49.7            \\
CMR-SG-RN50~\cite{chen2021camera}                  & -       & \cmark   & \xmark   & 48.8            \\
\rowcolor{lightergray} 
Baseline-Holisitc                                  & -       & \xmark   & \xmark   & 54.5             \\
\rowcolor{lightergray} 
Baseline-2D-Dense                                  & -       & \xmark   & \xmark   & 53.2             \\
\rowcolor{lightgray} 
CS-NVF (Ours)                                      & -       & \xmark   & \xmark   & \wincat{47.2}   \\
\hline
\rowcolor{lighterblue} 
Baseline-Holisitc                                  & -       & \xmark   & \cmark   & 50.4             \\
\rowcolor{lighterblue} 
Baseline-2D-Dense                                  & -       & \xmark   & \cmark   & 49.0             \\
\rowcolor{lightblue} 
CS-NVF (Ours)                                      & -       & \xmark   & \cmark   & \wincat{42.4}   \\
\hline
\rowcolor{lightgray} 
CS-NVF (Ours)                                      & Comp$^*$  & \xmark   & \xmark   & \wincat{44.6}   \\
\hline
\rowcolor{lightblue} 
CS-NVF (Ours)                                     & Comp$^*$  & \xmark   & \cmark   & \wincat{39.3}   \\
\hline
\end{tabular}
\vspace{-1ex}
\caption{\textbf{Comparison for absolute 3D hand pose on FreiHAND.} *: pre-training on Comp. Note that our CS-NVF and two baselines take the original image without hand detection or cropping.
}
\label{table:main_fh}
\end{minipage}
\hspace{6ex}
\begin{minipage}{\columnwidth}
\centering
\captionsetup{width=1.0\textwidth}
\begin{tabular}{ccccc}
\hline
\multirow{2}{*}{\vspace{2.5ex}Method} & {\begin{tabular}[c]{@{}c@{}}Extra \\ Data\end{tabular}} & {\begin{tabular}[c]{@{}c@{}}Hand \\ Scale\end{tabular}} & {TE}$\downarrow$ & {DE}$\downarrow$ \\ 
\hline
\rowcolor{lightergray} 
Baseline-Holisitc          & -       & \xmark   & 50.6             & 49.1 \\
\rowcolor{lightergray} 
Baseline-2D-Dense          & -       & \xmark   & 49.2             & 47.9 \\
\rowcolor{lightgray} 
CS-NVF (Ours)              & -       & \xmark   & \wincat{43.6}    & \wincat{42.4} \\
\hline
\rowcolor{lighterblue} 
Baseline-Holisitc          & -       & \cmark   & 46.9             & 45.5 \\
\rowcolor{lighterblue} 
Baseline-2D-Dense          & -       & \cmark   & 45.3             & 43.9 \\
\rowcolor{lightblue} 
CS-NVF (Ours)              & -       & \cmark   & \wincat{38.9}    & \wincat{37.8} \\ 
\hline
\rowcolor{lightergray} 
Baseline-Holisitc          & Comp$^*$  & \xmark   & 48.7             & 47.1 \\
\rowcolor{lightergray} 
Baseline-2D-Dense          & Comp$^*$  & \xmark   & 47.9             & 46.4 \\
\rowcolor{lightgray} 
CS-NVF (Ours)              & Comp$^*$  & \xmark   & \wincat{41.5}    & \wincat{40.4} \\
\hline
\rowcolor{lighterblue} 
Baseline-Holisitc          & Comp$^*$  & \cmark   & 41.7             & 40.1 \\
\rowcolor{lighterblue} 
Baseline-2D-Dense          & Comp$^*$  & \cmark   & 40.5             & 38.8 \\
\rowcolor{lightblue} 
CS-NVF (Ours)              & Comp$^*$  & \cmark   & \wincat{36.5}    & \wincat{35.5} \\ 
\hline
\end{tabular}
\vspace{-1ex}
\caption{\textbf{Comparison of 3D Translational and Depth Error for absolute 3D hand pose on FreiHAND.} *: pre-training on Comp.}
\label{table:depth_fh}
\end{minipage}
\vspace{-3ex}
\end{table*}
\customparagraph{Baseline-2D-Dense.}
We adopt a 2D dense regression scheme to estimate the camera-space pose via dense 2D pixel-wise voting, which is different from NVF which is based on dense 3D point-wise voting. Given the feature map $F^{''}$ extracted by the hourglass network $g(I;\boldsymbol{\eta^{''}})$, we only pass the pixel-aligned feature $F^{''}(\boldsymbol{u})$ at each 2D pixel location $\boldsymbol{u}$ through the MLP $f_\text{DEN}:\mathbb{R}^{C}\mapsto\mathbb{R} \times \mathbb{R}^{T\times4} \text{~as~}  f_\text{DEN}(F^{''}(\boldsymbol{u}); \boldsymbol{\theta^{''}}) = (e,V^{''})$. $f_\text{DEN}$ outputs the probability $e\in[0, 1]$ that the hand is present at $\boldsymbol{u}$ and also a set of 4D vectors $V^{''}=\{\boldsymbol{v}_t^{''}\}_{t=1}^{T}$.
Each 4D vector $\boldsymbol{v}_t^{''}=(w_t^{''},\boldsymbol{j}_t^{''})$ contains a 1D voting weight $w_t^{''}$ and 3D joint coordinate $\boldsymbol{j}_t^{''}$. For the 1D voting weight, it is defined similarly to the one in NVF. We first use the camera ray for each pixel to find its first 3D intersection point with the 3D hand model in the camera space. Then, we directly follow Eq.~\ref{eq:vote_weight} to calculate the pixel's voting weight as the closeness to each joint. 
It is trained by solving the following optimization problem with $\lambda^{''}$ as the balancing weight:
\begin{align}\label{eq:bd_loss}
    &\boldsymbol{\eta^{\prime\prime\star}}, \boldsymbol{\theta^{\prime\prime\star}} = \underset{\boldsymbol{\eta^{''},\theta^{''}}}{\arg\min} \;
    L_e + \lambda^{''} L_{V^{''}} \\
    &= \underset{\boldsymbol{\eta^{''},\theta^{''}}}{\arg\min} \;
     \frac{1}{L}\Big[\sum_{l=1}^L E\left(\hat{e}_l, {e}_l\right) + \lambda^{''}
     \sum_{l=1}^L \hat{e}_l H\left(\hat{V}_l^{''},V_l^{''}\right)\Big],
 \label{eq:loss_baseline}
\end{align}
where $L$ is the number of pixels, $E$ is the binary cross-entropy loss, $\hat{e}_l$ is given by the ground-truth hand mask, and $\hat{V}_l^{''}$ is the ground-truth set of 4D vectors. During inference, for pixel $\boldsymbol{u}_l$ with predicted probability $e_l>0.5$ (\ie, predicted as the foreground pixel), its predicted voting weight and 3D joint location are used to calculate the camera-space pose $J^{''} =\{\boldsymbol{j}_t^{''}\}_{t=1}^T$ by a weighted average, similar to Eq.~\ref{eq:pose}.

\subsection{Dataset and Evaluation Metrics}
\vspace{-1ex}
\customparagraph{FreiHAND Dataset~\cite{zimmermann2019freihand}.} 
As a real 3D hand dataset, it provides 130,240 training and 3,960 evaluation samples.
The result of the evaluation set is evaluated via the official server for online evaluation.
Since FreiHAND provides hand scale during evaluation time, we provide an ablation on hand scale for analysis of 2D-to-3D scale and depth ambiguity. 

\vspace{-1ex}
\customparagraph{Metrics.} For FreiHAND, we evaluate for the task of camera-space 3D hand pose estimation. As in~\cite{zimmermann2019freihand,chen2021camera}, we report the unaligned version of Mean Joint Error (mm) which is measured in camera space as \textbf{CS-MJE} and the Area-Under-the-Curve of PCK (Percentage of Correct Keypoints) vs. error thresholds as \textbf{CS-AUC} to evaluate the absolute 3D hand pose. We also define and report a 3D Translational Error (mm) \textbf{(TE)}~\cite{hodavn2016evaluation} and Depth Error (mm) \textbf{(DE)} to evaluate the predicted and ground-truth pose's centroid and the depth-axis of the centroid, respectively. The results of \textbf{TE} and \textbf{DE} will be used in analysis for baseline studies, especially for the analysis of scale and depth ambiguity.

\vspace{-1ex}
\customparagraph{HO3D Dataset~\cite{hampali2020honnotate}.} 
As a real 3D hand-object dataset, it contains 66,034 training samples and 11,524 evaluation samples.
We use the official server for online evaluation as the annotations of the evaluation set are not available.

\vspace{-1ex}
\customparagraph{Metrics.} For HO3D, we evaluate for the task of root-relative 3D hand pose estimation. As in~\cite{hampali2020honnotate,hampali2022keypoint}, we report the Mean Joint Error (mm) after scale-translation alignment of the root as \textbf{MJE} and corresponding \textbf{AUC} to evaluate the pose. We also follow~\cite{park2022handoccnet} to add ground-truth root to the predicted root-relative pose and report error (mm) as \textbf{RS-MJE} as it is equivalent to measuring in root-relative space.

\vspace{-1ex}
\customparagraph{Complement (Comp) Dataset~\cite{chen2022mobrecon}.} As a synthetic 3D hand dataset, it provides 1,520 poses and 71 viewpoints for research use. We only use it for pre-training when needed.

\subsection{Implementation Details}
\label{sec:imple}
For NVF and two baselines, the network architecture is adopted from PIFu~\cite{saito2019pifu}. 
Specifically, we use hourglass network with the output stride of $4\,$ and output channel of 256 as the image encoder.
All the MLPs $f_\text{NVF}$, $f_\text{HOL}$, and $f_\text{DEN}$ have four hidden fully-connected layers with the number of neurons as $(1024, 512, 256, 128)$.
Unless stated otherwise, during training for NVF, the clamping distance $\delta$ is set to $5$ mm. The ${K}$NN is set to $1024$. The ball radius $r$ is set to $80$ mm. During inference, the sampling step size of 3D points is $16$ mm in all three axes. The voting fraction is set to $50\%$.
Unlike PIFu, no image segmentation is applied. 
Camera intrinsic parameters are assumed known.
Base on the above setting, for camera-space 3D hand pose estimation, we directly sample 3D points in camera space, as described in Sec.~\ref{sec:nvf}. We call NVF working in camera space as CS-NVF. Unlike other methods in Tab.~\ref{table:main_fh}, CS-NVF and both baselines take the original image as input without hand detection or cropping.
For the task of root-relative 3D hand pose estimation, we follow similar sampling strategy but sample in a hand root-relative 3D cube instead of the camera frustum. We then call NVF working in root-relative space as RS-NVF. RS-NVF takes cropped hand-centered image. 
Please refer to the supplementary for more details.
\begin{figure}[t]
  \centering
	\includegraphics[width=1.0\linewidth]{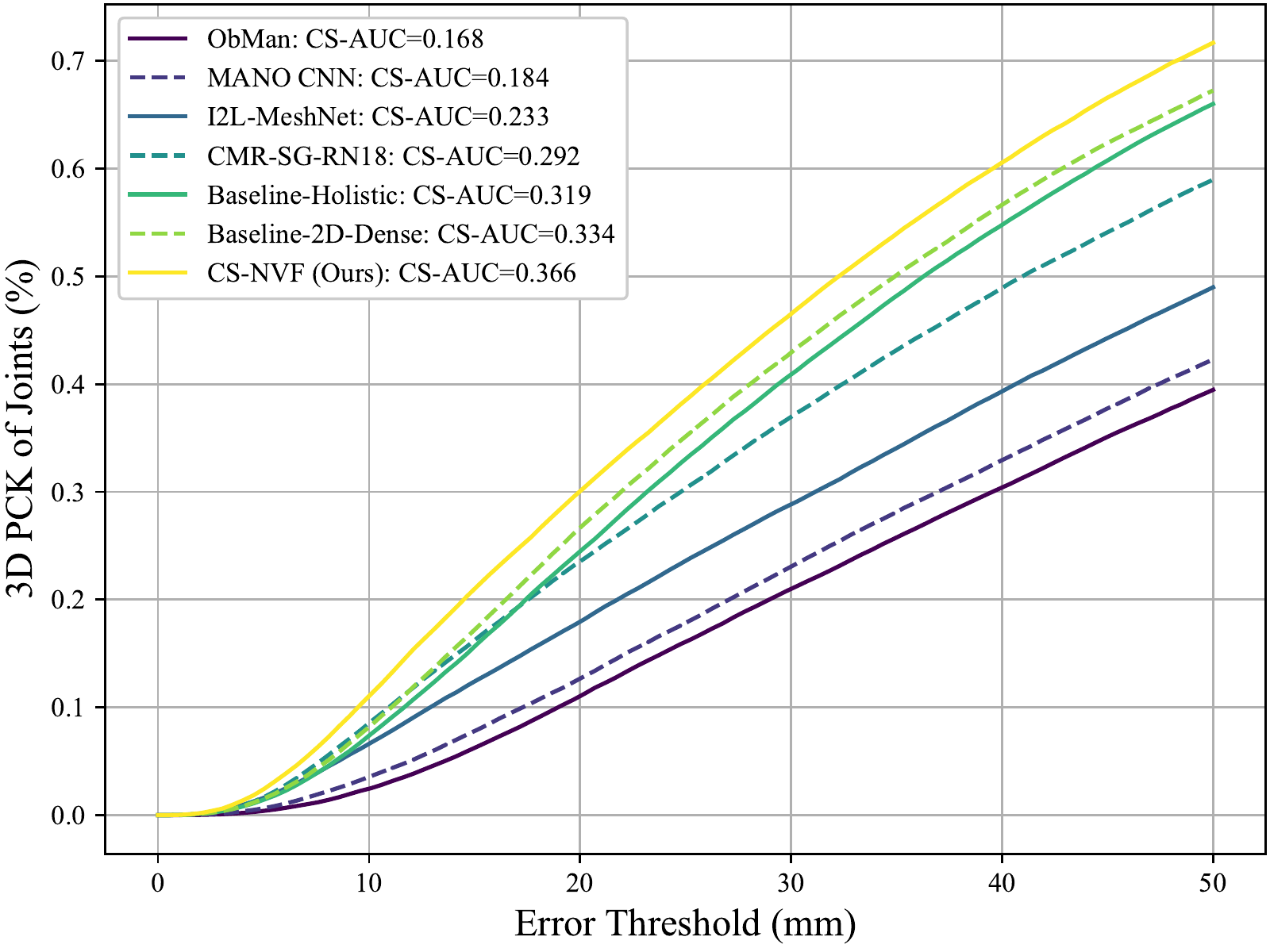}
    \vspace{-4ex}
   \caption{\textbf{3D PCK for absolute 3D hand pose on FreiHAND.}
   }
   \label{fig:pck}
\vspace{-3.5ex}
\end{figure}
\begin{figure*}[t]
	\begin{center}
         \hspace{0.3cm} {\footnotesize RGB input} \hspace{0.1cm} {\footnotesize Hand mask/mesh}  \hspace{0.5cm}  {\footnotesize Wrist} \hspace{0.6cm}   {\footnotesize Thumb root} \hspace{0.7cm}  {\footnotesize Index tip} \hspace{0.6cm}  {\footnotesize Middle root} \hspace{0.6cm}  {\footnotesize Ring tip} \hspace{0.6cm}  {\footnotesize Pinky root}  \hspace{0.3cm} {\footnotesize 3D hand pose}
	\vspace{-1.5ex}
    \includegraphics[width=1.0\linewidth]{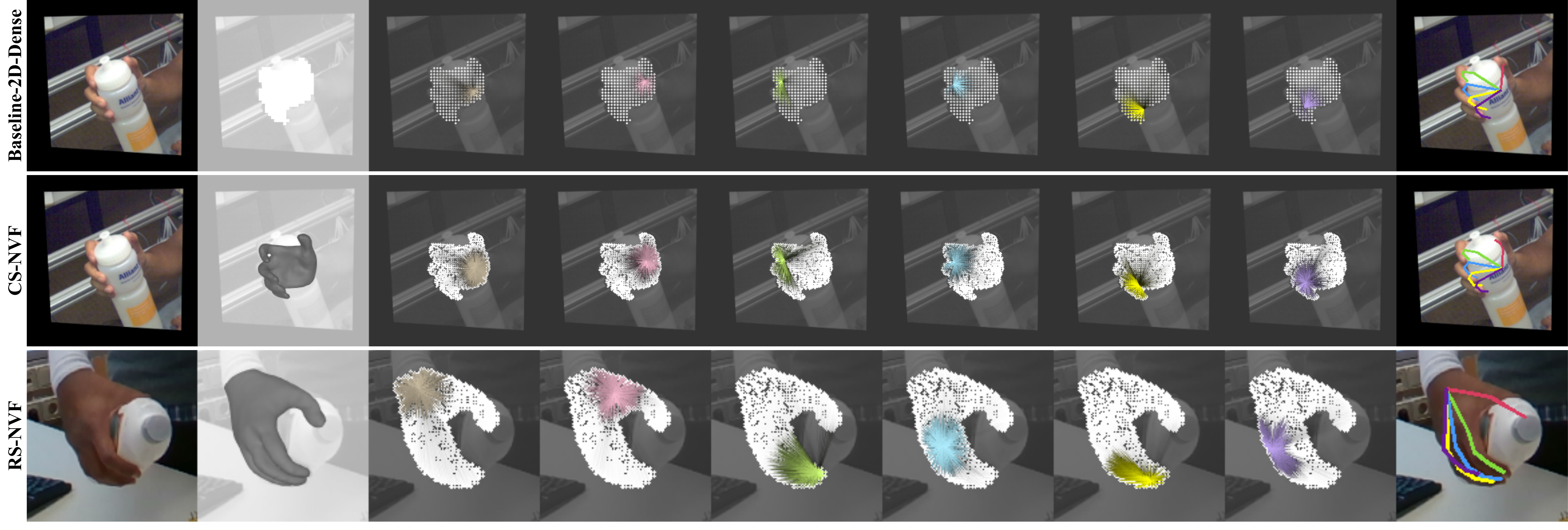}
	\vspace{-1.5ex}
        \hspace{0.2cm} {\footnotesize (a)} \hspace{1.45cm}   {\footnotesize (b)}  \hspace{1.5cm}  {\footnotesize (c)} \hspace{1.4cm}  {\footnotesize (d)} \hspace{1.4cm}  {\footnotesize (e)} \hspace{1.5cm}  {\footnotesize (f)} \hspace{1.4cm}  {\footnotesize (g)} \hspace{1.3cm}  {\footnotesize (h)} \hspace{1.5cm}  {\footnotesize (i)}
    	\caption{\label{fig:combine_viz} \textbf{Qualitative results for absolute and relative 3D pose:}
    	(b) Predicted hand mask (Baseline) or mesh generated by Marching Cubes from the predicted signed distances (NVF).
        (c-h) Predicted 1D voting weight from each predicted foreground pixel (Baseline) or predicted near-surface 3D point (NVF) to a joint. (Brighter line means larger weight, darker or no line means smaller weight)
        (i) Estimated 3D pose. NVF makes solid point-wise prediction and overall distribution of signed distances and voting weights in highly occluded region.}
	\end{center}
	\vspace{-5ex}
\end{figure*}
\vspace{-1ex}
\subsection{Baseline Studies}
\vspace{-1ex}
\label{sec:base_study}
In this section, based on the results from CS-NVF and both baselines (as in Tab.~\ref{table:main_fh}, Tab.~\ref{table:depth_fh}, Fig.~\ref{fig:pck}, and Fig.~\ref{fig:combine_viz}), we investigate the impact of 2D-to-3D scale and depth ambiguity for the task of absolute 3D hand pose estimation and the effectiveness of using 3D dense regression over both 2D dense regression and holistic regression.

\vspace{-1ex}
\customparagraph{2D-to-3D Scale and Depth Ambiguity.}
Estimating absolute 3D pose from an RGB image is an ill-posed problem due to 2D-to-3D scale and depth ambiguity ~\cite{dijk2019neural,iqbal2018hand}. In view of this mater, FreiHAND~\cite{zimmermann2019freihand} provides hand scale during evaluation for participants to use. 
As most existing works do not use the provided hand scale and thus, no relevant ablation on hand scale exists, we provide ablation results on hand scale in Tab.~\ref{table:main_fh} and Tab.~\ref{table:depth_fh} for CS-NVF and two baselines. First, based on results in Tab.~\ref{table:main_fh} and Tab.~\ref{table:depth_fh}, we observe that, for all three pipelines, the main cause of the error between the ground-truth and predicted hand joints is the global translation and especially the depth component as TE is close to CS-MJE and DE is close to TE.
For instance, CS-NVF with hand scale and Comp pre-training achieves $39.3$ mm CS-MJE, while TE and DE are $36.5$ mm and $35.5$ mm, respectively. 
This validates the fact that depth ambiguity is indeed one of the main challenges for estimating accurate absolute 3D hand pose from a single RGB image.
Moreover, for CS-NVF and the two baselines, using hand scale leads to significant improvements in terms of all three metrics. 
It is also worth noting that, for all three pipelines, results only using hand scale outperform results using only extra data. For instance, CS-NVF with hand scale achieves $42.4$ mm CS-MJE, while CS-NVF with Comp pre-training has $44.6$ mm CJ-MJE.
These verify that a reasonable understanding towards the hand scale would help for robust depth learning and alleviate depth ambiguity, which is also aligned with existing literature~\cite{dijk2019neural,iqbal2018hand,lepetit2009epnp}.

\vspace{-1ex}
\customparagraph{Dense Regression vs. Holistic Regression.}
In Tab.~\ref{table:main_fh} and Fig.~\ref{fig:pck}, we can observe that both CS-NVF and Baseline-2D-Dense consistently outperform Baseline-Holistic on CS-MJE and CS-AUC under different training settings. 
For instance, without using hand scale, Baseline-2D-Dense is $1.3$ mm better than Baseline-Holistic in CS-MJE, and CS-NVF can be $7.3$ mm better.  
Since three pipelines share the same network architecture, this supports our point that dense regression is more effective than holistic regression for handling highly articulated pose data.

\begin{table}[t]
\centering
\begin{tabular}{cccc}
\hline
Method & MJE$\downarrow$ & AUC$\uparrow$ & RS-MJE$\downarrow$\\ 
\hline
Hasson et al.~\cite{hasson2020leveraging}            & 36.9            & 0.369            & - \\
Pose2Mesh~\cite{choi2020pose2mesh}                   & 33.3            & 0.480            & 33.2 \\
ObMan~\cite{hasson19_obman}                          & 31.8            & 0.461            & 55.2 \\
Liu et al.~\cite{liu2021semi}                        & 31.7            & 0.463            & 30.0 \\
Hampali et al.~\cite{hampali2020honnotate}           & 30.4            & 0.494            & - \\
METRO~\cite{lin2021end}                              & 28.9            & 0.504            & - \\ 
I2L-MeshNet~\cite{moon2020i2l}                       & 26.0            & 0.529            & 26.8 \\
Keypoint Trans.~\cite{hampali2022keypoint}           & 25.7            & 0.553            & - \\
ArtiBoost~\cite{yang2022artiboost}                   & 25.3            & 0.532            & - \\
Zheng et al.~\cite{zheng2022rethink}                 & 25.1            & 0.541            & - \\
HandOccNet~\cite{park2022handoccnet}                 & 24.0            & 0.557            & 24.9 \\
\rowcolor{lightgray} 
RS-NVF (Ours)            & \textbf{21.8}   & \wincat{0.610}   & \wincat{23.2} \\ 
\hline
\end{tabular}
\vspace{-1ex}
\caption{\textbf{Comparison for relative 3D hand pose on HO3D.}
}
\label{table:main_ho}
\vspace{-3.5ex}
\end{table}
\begin{table*}[t]
\centering
\begin{tabular}{ccccccccccccccccc}
\hline
\multirow{2}{*}{\vspace{3ex}Proposed} & {Evaluation} &  & \multicolumn{4}{c}{Voting Fraction (\%)} &  & \multicolumn{4}{c}{Clamping Distance (mm)} &  & \multicolumn{4}{c}{$K$NN} \\ \cline{4-7} \cline{9-12} \cline{14-17} 
{Method}   & {Metric}                    & & 25 & 50 & 75 & 100        &   & 1   & 5  & 10   & 20  &   & 256 & 512 & 1024 & 2048 \\ \cline{1-2} \cline{4-7} \cline{9-12} \cline{14-17} 
\multicolumn{1}{c}{\multirow{2}{*}{\vspace{2.5ex}CS-NVF}} & CS-MJE$\downarrow$   & & 39.9   & 39.3   & 39.5   & 40.2   &   & 39.3   & 39.3   & 40.0   & 40.1   &   & 40.3   & 39.5   & 39.3   & 39.3 \\
\multicolumn{1}{c}{{(Ours)}} & CS-AUC$\uparrow$     & & 0.36   & 0.37   & 0.36   & 0.36   &   & 0.37   & 0.37   & 0.36   & 0.36   &   & 0.35   & 0.36   & 0.37   & 0.37 \\ \cline{1-2} \cline{4-7} \cline{9-12} \cline{14-17} 
\multicolumn{1}{c}{\multirow{2}{*}{\vspace{2.5ex}RS-NVF}} & MJE$\downarrow$      & & 22.2   & 21.8   & 21.9   & 22.0   &   & 21.9   & 21.8   & 21.8   & 21.9   &   & 23.6   & 22.6   & 21.8   & 21.8 \\
\multicolumn{1}{c}{{(Ours)}} & AUC$\uparrow$        & & 0.60   & 0.61   & 0.61   & 0.60   &   & 0.60   & 0.61   & 0.61   & 0.61   &   & 0.59   & 0.60   & 0.61   & 0.61 \\ \hline
\end{tabular}
\vspace{-1ex}
\caption{\textbf{Ablation study on parameter choices for absolute and relative 3D hand pose:} Voting fraction, clamping distance, and $K$NN
}
\label{table:ab_param}
\vspace{-3ex}
\end{table*}
\begin{table}[t]
\centering
\begin{tabular}{cccccc}
\hline
Step &  & \multicolumn{4}{c}{CS-NVF (Ours)} \\ \cline{3-6} 
Size &  & \#Points & CS-MJE$\downarrow$  & CS-AUC$\uparrow$  & FPS$\uparrow$  \\ \cline{1-1} \cline{3-6} 
4    &  & 1882k  & 39.2   & 0.37    & 1   \\
8    &  & 230k   & 39.2   & 0.37    & 8    \\
\rowcolor{lightblue} 
16   &  & 28k    & 39.3   & 0.37    & 34   \\
24   &  & 8k     & 39.8   & 0.36    & 50   \\ 
32   &  & 3k     & 42.4   & 0.34    & 56   \\ \hline
\end{tabular}
\vspace{-1ex}
\caption{\textbf{Ablation study on sampling step size for absolute 3D hand pose on FreiHAND.} \#points: number of sampled 3D points.  
}
\label{table:ab_step}
\vspace{-3ex}
\end{table}
\vspace{-1ex}
\customparagraph{3D Dense Regression vs. 2D Dense Regression.}
Quantitatively, in Tab.~\ref{table:main_fh} and Fig.~\ref{fig:pck}, it can be observed that CS-NVF consistently outperforms Baseline-2D-Dense by a large margin on CS-MJE and CS-AUC under different training settings.
For instance, without using hand scale, CS-NVF can be $6$ mm better on CS-MJE.
Qualitatively, as indicated by the white circles in column (c-h) of Fig.~\ref{fig:combine_viz}, both methods are able to find valid voters (Baseline-2D-Dense: pixels with predicted probability $e>0.5$. CS-NVF: 3D points with predicted signed distance $|s|<\delta$).
However, CS-NVF generally provides not only more number of valid voters, but also more reasonable element-wise prediction and overall distribution of the voting weight than Baseline-2D-Dense.  
Specifically, for the number of voters, since CS-NVF models all the 3D points around the hand surface and  Baseline-2D-Dense can only model foreground pixels, CS-NVF generally can find and use more valid voters.
As for element-wise prediction and overall distribution of the voting weight, CS-NVF is generally able to predict large voting weight for points close to hand joint and small value for points far away, resulting in well-shaped symmetric distribution around the corresponding joint, even in occluded regions. Note that brighter line indicates larger voting weight, darker or no line indicates smaller voting weight. 
Moreover, as shown in column (b) for the mesh rendering of CS-NVF (obtained using Marching Cubes~\cite{lorensen1987marching} from predicted signed distances), even in highly occluded region, CS-NVF provides solid signed distance distribution showing its ability to reason the global hand geometry.
Overall, these demonstrate that CS-NVF, through direct dense modeling in 3D domain, can better model 3D dense local evidence, the relation between voters and 3D hand pose, and also the global hand geometry, compared with 2D dense regression-based counterpart, which leads to robust 3D hand pose estimation.

\vspace{-0.5ex}
\subsection{Main Results: Absolute 3D Hand Pose}
\vspace{-0.5ex}
In Tab.~\ref{table:main_fh} and Fig.~\ref{fig:pck}, we also compare CS-NVF with current state-of-the-art methods on FreiHAND for absolute 3D hand pose estimation. In Tab.~\ref{table:main_fh}, without the use of hand scale and extra data as a fair comparison, CS-NVF surpasses previous state-of-the-art methods, achieving $1.6$ mm better than CMR~\cite{chen2021camera}. Note that, unlike other methods, CS-NVF takes the original image as input without hand detection or cropping.
Additional hand scale information used without extra data improves the CS-MJE of CS-NVF further to $42.4$ mm.
Additional Comp pre-training improves the CS-MJE of CS-NVF to $44.6$ mm without hand scale and $39.3$ mm with hand scale. 
Referring to Fig.~\ref{fig:pck}, CS-NVF also achieves state-of-the-art performance on CS-AUC.

\vspace{-0.5ex}
\subsection{Main Results: Relative 3D Hand Pose}
\vspace{-0.5ex}
In Tab.~\ref{table:main_ho}, we compare RS-NVF with current state-of-the-art methods on HO3D for relative 3D hand pose estimation. RS-NVF significantly outperforms all these methods in all three metrics. Especially, RS-NVF outmatches HandOccNet~\cite{park2022handoccnet} by $2.2$ mm on MJE, $1.7$ mm on RS-MJE, and also has $0.610$ AUC agaist $0.557$ AUC from HandOccNet.

\vspace{-0.5ex}
\subsection{Ablation Studies}
\label{sec:ablation}
\vspace{-1.5ex}
\customparagraph{Sampling Step Size.}
During inference, 3D query points are sampled at centers of voxels that fill up the 3D space. Since the choices of different sampling step sizes can influence both accuracy and efficiency, in Tab.~\ref{table:ab_step}, we show the ablation on sampling step size for CS-NVF.
With $16$ mm as the step size, CS-NVF achieves $34$ FPS and $39.3$ mm on CS-MJE given 28k points.
Reducing the step size to $4$ mm yields 1,882k points and 1 FPS with slightly improved CS-MJE of $39.2$ mm.
Increasing to $32$ mm step size yields 3k points and improves the FPS to $56$ with still competitive $42.4$ mm on CS-MJE.
For RS-NVF not shown in the table, $4$ and $8$ mm step size both yield MJE of $21.7$ mm and AUC of $0.62$.
We can observe that, while step size affects FPS, NVF is relatively insensitive regarding the performance.
In all other reported results, we use $16$ mm as the default step size.

\vspace{-1ex}
\customparagraph{Voting Fraction.}
Since the predicted point-wise voting weight reflects the confidence and closeness of a given point towards a certain joint, besides using all the near-surface points as valid voters, we can use a specified fraction of near-surface points with the highest voting weight for voting during inference. In Tab.~\ref{table:ab_param}, we observe that using all the near-surface points ($100\%$) yields competitive results while $50\%$ fraction leads to the overall better performance for both CS-NVF and RS-NVF. 
The results are aligned with our algorithm design since points far from each joint are not expected to make meaningful contributions.
In all other reported results, we use $50\%$ as the default voting fraction.

\vspace{-1ex}
\customparagraph{Clamping Distance.}
During training, clamping distance $\delta$ is used to concentrate NVF's capacity on learning high-quality point-wise prediction (\ie, signed distance and set of 4D offset vectors) near the hand surface. Ablation in Tab.~\ref{table:ab_param} shows that $5$ mm generates the overall better performance for both CS-NVF and RS-NVF.
In all other reported results, we use $5$ mm as the default clamping distance.

\vspace{-1ex}
\customparagraph{$K$-Nearest Neighbors (${K}$NN).}
During training, among all the sampled points that are in the hand surface vicinity, non-zero 4D offset vector will only be assigned to the $K$ nearest points inside a 3D ball of each hand joint. 
As shown in previous works~\cite{wan2018dense,ge2018point}, estimating offsets for large $K$NN is unnecessary and far away joints are beyond the scope of receptive field. Small $K$NN might also cause poor performance due to the sparse learning scheme. Ablation in Tab.~\ref{table:ab_param} shows that 1024 $K$NN generates the overall better performance for both CS-NVF and RS-NVF.
In all other reported results, we use $1024$ as the default $K$NN.
\section{Conclusions}
\label{sec:conclusions}
We propose Neural Voting Field (NVF), as the first 3D implicit representation-based unified solution to estimate camera-space 3D hand pose given a single RGB image.
NVF follows a novel unified 3D dense regression scheme to estimate 3D hand pose in 3D camera space via dense 3D point-wise voting in camera frustum.
Studies with both holistic regression baseline and 2D dense regression baseline verify NVF's ability to fully exploit 3D dense local evidence and model global hand structure/geometry, which are essential for precisely locating 3D hand joints in 3D camera space. 
NVF also demonstrates state-of-the-art results on both tasks of camera-space 3D hand pose estimation and root-relative 3D hand pose estimation.

\clearpage

{\small
\bibliographystyle{ieee_fullname}
\bibliography{references}
}

\clearpage

\renewcommand\thesection{\Alph{section}}
\renewcommand\thesubsection{\thesection.\Alph{subsection}}
\setcounter{section}{0}
\setcounter{table}{6}
\setcounter{figure}{4}

\appendix
\pdfoutput=1
\twocolumn[{
\renewcommand\twocolumn[1][]{#1}
\begin{center}
\textbf{\Large Neural Voting Field for Camera-Space 3D Hand Pose Estimation\\{Supplementary Material}}
\end{center}
\vspace{4mm}
}]

This is the supplementary materials of the main text. 
Sec.~\ref{sec:supp_imple} provides more details regarding the method configurations and training procedure. 
Sec.~\ref{sec:fh_table} presents full quantitative results of the two baselines (\ie, Baseline-Holistic and Baseline-2D-Dense) for camera-space 3D hand pose on FreiHAND~\cite{zimmermann2019freihand} with Comp~\cite{chen2022mobrecon} pre-training.
Sec.~\ref{sec:comparison} presents qualitative comparisons between our CS-NVF and the state-of-the-art methods (\ie, CMR~\cite{chen2021camera} and I2L-MeshNet~\cite{moon2020i2l}) for the task of camera-space 3D hand pose estimation on complex and failure cases (\eg, severe occlusion and extreme poses) on FreiHAND~\cite{zimmermann2019freihand}.
Sec.~\ref{sec:qualitative} provides additional qualitative results from CS-NVF for camera-space 3D hand pose on FreiHAND~\cite{zimmermann2019freihand} and RS-NVF for root-relative 3D hand pose on HO3D~\cite{hampali2020honnotate}.
To demonstrate the generalization ability of NVF, Sec.~\ref{sec:gen} provides qualitative results on Real-World dataset~\cite{ge20193d} using CS-NVF trained on FreiHAND only. 

\vspace{-1ex}
\section{Additional Implementation Details}
\vspace{-0.5ex}
\label{sec:supp_imple}
Continuing from Sec.~\ref{sec:imple} of implementation details in the main text, we provide more details regarding the method configurations and training procedure in this section.

\vspace{-0.5ex}
\customparagraph{Additional Method Configurations.}
For each single RGB image used for all of our models, different from PIFu~\cite{saito2019pifu}, no image segmentation is applied. 
Moreover, for camera-space 3D hand pose estimation, CS-NVF and both baselines take the original image with resolution $224\times224$ as input without hand detection or cropping applied.
As in~\cite{huang2022neural}, to remove the ambiguity caused by using images captured by cameras with different focal lengths during training for absolute 3D hand pose estimation, given the provided camera intrinsic parameters from FreiHAND, we remap each input image to a reference pinhole camera with the same focal length which can be arbitrarily chosen. 
For root-relative 3D hand pose estimation, RS-NVF takes cropped hand-centered image with resolution $128\times128$ as input without remapping applied.
During training, the balancing weight used for CS-NVF and Baseline-2D-Dense is set to $0.1$ and the balancing weight used for RS-NVF is set to $10$.

\customparagraph{Training Procedure.}
We report results achieved by the proposed NVF and the two baselines under various training settings (\ie, with and without hand scale or extra data). 
For our models without the use of hand scale and extra data as a fair comparison with the state-of-the-art methods, the image encoder ($g$) and MLP ($f_\text{NVF}$, $f_\text{HOL}$, $f_\text{DEN}$) are initialized via xavier initialization.
RMSProp~\cite{tieleman2012rmsprop} is used for optimization with the batch size of 24, the number of epochs of 650, and the initial learning rate of $0.0001$. The learning rate is decayed by the factor of $0.1$ at $400$-th, $500$-th, and $600$-th epoch.
The models can also be pre-trained first on FreiHAND to estimate relative 3D hand pose. 
RMSProp is used for optimization with the batch size of 64, the number of epochs of 600, and the initial learning rate of 0.0005. The learning rate is decayed by the factor of 0.1 at $250$-th, $350$-th, and $450$-th epoch.
We then fine-tune the models for respective tasks. 
RMSProp is used for optimization with the batch size of 24, the number of epochs of 60, and the initial learning rate of 0.0001. The learning rate is decayed by the factor of 0.1 at $25$-th and $50$-th epoch.
We found that pre-training on FreiHAND to estimate relative hand pose helps to improve generalization on input images from other domains, especially for unseen poses.
For results using extra data, 
the models are pre-trained first on Comp to estimate relative hand pose and 
RMSProp is used for optimization with the batch size of 64, the number of epochs of $700$, and the initial learning rate of $0.0005$. The learning rate is decayed by the factor of $0.1$ at $500$-th, and $600$-th epoch.  
For our ablation study on hand scale for camera-space 3D hand pose, we directly use the hand scale provided by FreiHAND during evaluation. The hand scale is defined as the metric length of a reference bone which is the phalangal proximal bone of the middle finger. When hand scale is used, it is concatenated with the input to the MLP for processing.

\begin{table*}[t]
\begin{minipage}{\columnwidth}
\centering
\captionsetup{width=1.0\textwidth}
\begin{tabular}{cccccc}
\hline
\multirow{2}{*}{\vspace{2.5ex}Method} & {\begin{tabular}[c]{@{}c@{}}Extra \\ Data\end{tabular}} & {\begin{tabular}[c]{@{}c@{}}Hand \\ Crop\end{tabular}} & {\begin{tabular}[c]{@{}c@{}}Hand \\ Scale\end{tabular}} & {CS-MJE}$\downarrow$ \\ 
\hline
ObMan~\cite{hasson19_obman}                        & -       & \cmark   & \xmark   & 85.2             \\
MANO CNN~\cite{zimmermann2019freihand}             & -       & \cmark   & \xmark   & 71.3             \\
I2L-MeshNet~\cite{moon2020i2l}                     & -       & \cmark   & \xmark   & 60.3            \\
CMR-SG-RN18~\cite{chen2021camera}                  & -       & \cmark   & \xmark   & 49.7            \\
CMR-SG-RN50~\cite{chen2021camera}                  & -       & \cmark   & \xmark   & 48.8            \\
\rowcolor{lightergray} 
Baseline-Holisitc                                  & -       & \xmark   & \xmark   & 54.5             \\
\rowcolor{lightergray} 
Baseline-2D-Dense                                  & -       & \xmark   & \xmark   & 53.2             \\
\rowcolor{lightgray} 
CS-NVF (Ours)                                      & -       & \xmark   & \xmark   & \wincat{47.2}   \\
\hline
\rowcolor{lighterblue} 
Baseline-Holisitc                                  & -       & \xmark   & \cmark   & 50.4             \\
\rowcolor{lighterblue} 
Baseline-2D-Dense                                  & -       & \xmark   & \cmark   & 49.0             \\
\rowcolor{lightblue} 
CS-NVF (Ours)                                      & -       & \xmark   & \cmark   & \wincat{42.4}   \\
\hline
\rowcolor{lightergray} 
Baseline-Holisitc                               & Comp$^*$  & \xmark   & \xmark   & 51.3              \\
\rowcolor{lightergray} 
Baseline-2D-Dense                               & Comp$^*$  & \xmark   & \xmark   & 50.9             \\
\rowcolor{lightgray} 
CS-NVF (Ours)                                   & Comp$^*$  & \xmark   & \xmark   & \wincat{44.6}   \\
\hline
\rowcolor{lighterblue} 
Baseline-Holisitc                               & Comp$^*$  & \xmark   & \cmark   & 44.3            \\
\rowcolor{lighterblue} 
Baseline-2D-Dense                               & Comp$^*$  & \xmark   & \cmark   & 43.4             \\
\rowcolor{lightblue} 
CS-NVF (Ours)                                   & Comp$^*$  & \xmark   & \cmark   & \wincat{39.3}   \\
\hline
\end{tabular}
\vspace{-1ex}
\caption{\textbf{Comparison for absolute 3D hand pose on FreiHAND.} *: pre-training on Comp. Note that our CS-NVF and two baselines take the original image without hand detection or cropping.
}
\label{table:supp_main_fh}
\end{minipage}
\hspace{6ex}
\begin{minipage}{\columnwidth}
\centering
\captionsetup{width=1.0\textwidth}
\begin{tabular}{ccccc}
\hline
\multirow{2}{*}{\vspace{2.5ex}Method} & {\begin{tabular}[c]{@{}c@{}}Extra \\ Data\end{tabular}} & {\begin{tabular}[c]{@{}c@{}}Hand \\ Scale\end{tabular}} & {TE}$\downarrow$ & {DE}$\downarrow$ \\ 
\hline
\rowcolor{lightergray} 
Baseline-Holisitc          & -       & \xmark   & 50.6             & 49.1 \\
\rowcolor{lightergray} 
Baseline-2D-Dense          & -       & \xmark   & 49.2             & 47.9 \\
\rowcolor{lightgray} 
CS-NVF (Ours)              & -       & \xmark   & \wincat{43.6}    & \wincat{42.4} \\
\hline
\rowcolor{lighterblue} 
Baseline-Holisitc          & -       & \cmark   & 46.9             & 45.5 \\
\rowcolor{lighterblue} 
Baseline-2D-Dense          & -       & \cmark   & 45.3             & 43.9 \\
\rowcolor{lightblue} 
CS-NVF (Ours)              & -       & \cmark   & \wincat{38.9}    & \wincat{37.8} \\ 
\hline
\rowcolor{lightergray} 
Baseline-Holisitc          & Comp$^*$  & \xmark   & 48.7             & 47.1 \\
\rowcolor{lightergray} 
Baseline-2D-Dense          & Comp$^*$  & \xmark   & 47.9             & 46.4 \\
\rowcolor{lightgray} 
CS-NVF (Ours)              & Comp$^*$  & \xmark   & \wincat{41.5}    & \wincat{40.4} \\
\hline
\rowcolor{lighterblue} 
Baseline-Holisitc          & Comp$^*$  & \cmark   & 41.7             & 40.1 \\
\rowcolor{lighterblue} 
Baseline-2D-Dense          & Comp$^*$  & \cmark   & 40.5             & 38.8 \\
\rowcolor{lightblue} 
CS-NVF (Ours)              & Comp$^*$  & \cmark   & \wincat{36.5}    & \wincat{35.5} \\ 
\hline
\end{tabular}
\vspace{-1ex}
\caption{\textbf{Comparison of 3D Translational and Depth Error for absolute 3D hand pose on FreiHAND.} *: pre-training on Comp.}
\label{table:supp_depth_fh}
\end{minipage}
\vspace{-3ex}
\end{table*}
\vspace{-1ex}
\section{Baseline Results with Extra Data}
\vspace{-0.5ex}
\label{sec:fh_table}
Besides the results of the two baselines shown in Tab.~\ref{table:main_fh} and Tab.~\ref{table:depth_fh} of the main paper for camera-space 3D hand pose estimation, we also provide results of the two baselines on the metric of CS-MJE with Comp pre-training in Tab.~\ref{table:supp_main_fh}. This then provides the full results from CS-NVF and the two baselines under different training settings (\ie, with and without hand scale or extra data) for camera-space 3D hand pose on FreiHAND, as shown in Tab.~\ref{table:supp_main_fh} and Tab.~\ref{table:supp_depth_fh}.

\vspace{-1ex}
\section{Qualitative Comparisons on Complex and Failure Cases}
\begin{figure*}[t]
	\begin{center}
    \includegraphics[width=1.0\linewidth]{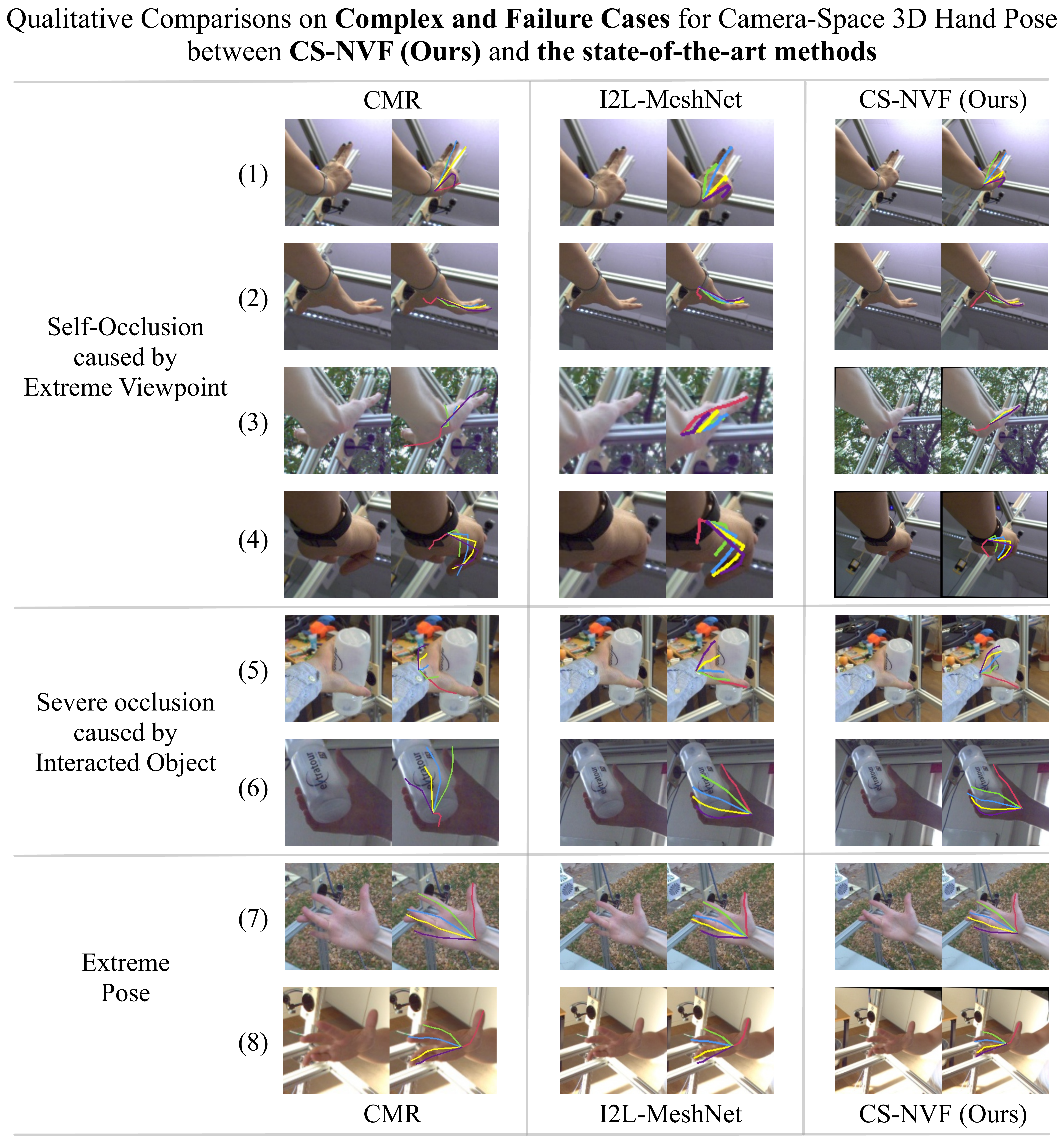}
	\end{center}
	\caption{\label{fig:hard} \textbf{Qualitative comparisons on complex and failure cases for absolute 3D hand pose on FreiHAND.} 
	For each pair of images, the left shows the input RGB image that the method uses during inference and the right shows the predicted camera-space 3D hand pose directly rendered by camera intrinsic parameters.
	Compared with the state-of-the-art methods (\ie, CMR~\cite{chen2021camera} and I2L-MeshNet~\cite{moon2020i2l}) on complex and failure cases
    with respect to the pose plausibility, pose accuracy, and 3D-2D alignment
	, CS-NVF has shown its capability to recover robust 3D hand pose in camera space when facing severe self-occlusion casued by extreme viewpoint, occlusion caused by interacted object, and extreme pose. 
	}
\end{figure*}

\vspace{-1ex}
\label{sec:comparison}
In Fig.~\ref{fig:hard}, we provide qualitative comparisons between our proposed CS-NVF and the state-of-the-art methods (\ie, CMR~\cite{chen2021camera} and I2L-MeshNet~\cite{moon2020i2l}) for the task of camera-space 3D hand pose estimation on complex and failure cases on FreiHAND~\cite{zimmermann2019freihand}. 
Specifically, the complex and failure cases as shown in Fig.~\ref{fig:hard} can be divided into three categories:

\vspace{-1ex}
\begin{itemize}
\setlength\itemsep{0em}
\item Severe self-occlusion caused by extreme viewpoint.
\item Severe occlusion caused by hand-object interaction. 
\item Extreme pose. 
\end{itemize}
\vspace{-1ex}
For each pair of images, the left shows the input RGB image that the method uses during inference and the right shows the predicted camera-space 3D hand pose directly rendered by camera intrinsic parameters.
Note that, as shown in the figure, CS-NVF takes the original RGB image as input without hand detection and cropping, while both CMR and I2L-MeshNet take the cropped image.
Based on the qualitative comparisons shown in Fig.~\ref{fig:hard} from three methods, we observe that:
\begin{itemize}
\setlength\itemsep{0em}
\vspace{-1ex}
\item Self-occlusion: for results shown in row (1-4) of Fig.~\ref{fig:hard}, when only a small portion of the hand is visible in the input image caused by extreme viewpoint, our proposed CS-NVF robustly recovers more plausible and accurate pose structure than CMR and I2L-MeshNet.
For example, in row (1), CMR shows three fingers up given the input hand with two fingers up and in row (3), I2L-MeshNet generates implausible structure.
\item Object occlusion: for results shown in row (5-6) of Fig.~\ref{fig:hard}, with severe occlusion caused by the interacted bottle, CS-NVF can better recover the occluded four fingers behind the bottle and show an overall reasonable gesture of hand grabbing a bottle. 
In row (6), CS-NVF is able to provide solid estimation for index finger which is entirely occluded. 
\vspace{-1ex}
\item Extreme pose: for results shown in row (7-8) of Fig.~\ref{fig:hard}, while all three methods can generate plausible hand pose structure, the three methods all fail at the part where the two fingers are crossed. 
Challenging poses shown in the row (7-8) usually are tail-distributed poses in most hand datasets. Thus, to tackle this problem, we argue that improving both fine-grained reasoning towards uncommon gestures for pose estimation pipeline and the pose distribution for hand dataset are required. Both aspects will be investigated in our future work.
\vspace{-1ex}
\item 3D-2D alignment: Besides the plausibility and accuracy of pose articulated structure itself, for the task of camera-space 3D hand pose estimation, we also need to look at the 3D global information (\ie, rotation and translation), which can be indicated by the alignment between the rendered hand pose and input hand area to some extent. With severe occlusion as shown in row (1-6), rendered 3D poses from CS-NVF generally show better 3D-2D alignment.
\vspace{-1ex}
\item Overall: while the performances on these challenging scenarios from all three methods are usually worse than performances on common cases, CS-NVF has shown its capability to recover more robust 3D hand pose in camera space in various challenging scenarios, compared with the state-of-the-art methods (i.e., CMR and I2L-MeshNet) with respect to the pose plausibility, pose accuracy, and 3D-2D alignment. 
\end{itemize}

\vspace{-2ex}
\section{Additional Qualitative Results}
\label{sec:qualitative}
\begin{figure*}[t]
	\begin{center}
         \hspace{0.0cm} {\footnotesize RGB input} \hspace{0.3cm} {\footnotesize 3D hand mesh}  \hspace{0.7cm}  {\footnotesize Wrist} \hspace{0.7cm}   {\footnotesize Thumb root} \hspace{0.5cm}  {\footnotesize Index tip} \hspace{0.7cm}  {\footnotesize Middle root} \hspace{0.6cm}  {\footnotesize Ring tip} \hspace{0.8cm}  {\footnotesize Pinky root}  \hspace{0.35cm} {\footnotesize 3D hand pose}
    \includegraphics[width=1.0\linewidth]{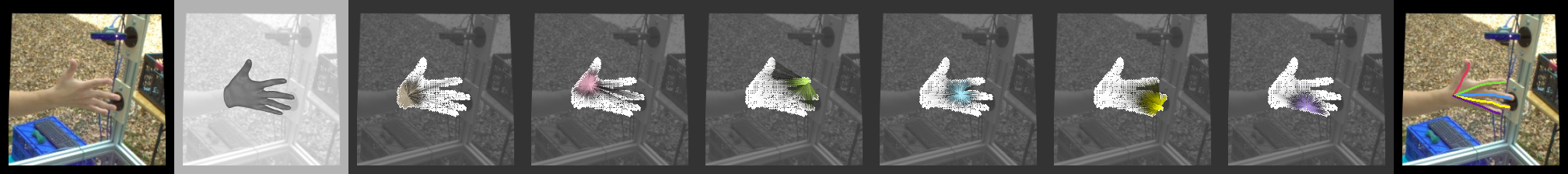}
    \includegraphics[width=1.0\linewidth]{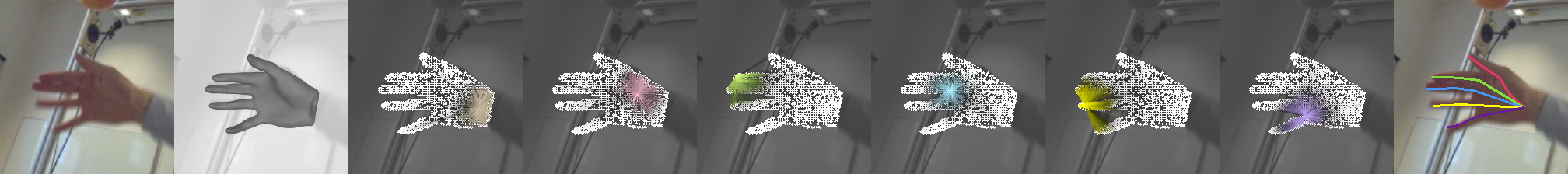}
    \includegraphics[width=1.0\linewidth]{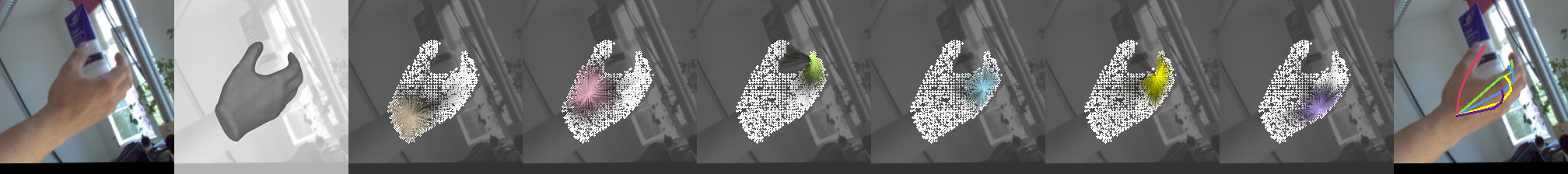}
    \includegraphics[width=1.0\linewidth]{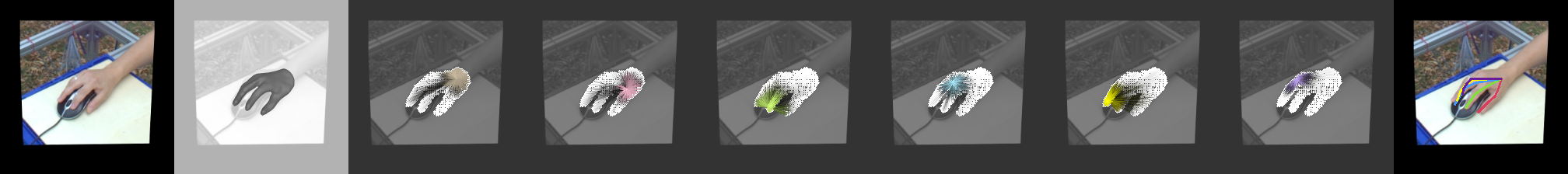}
    \includegraphics[width=1.0\linewidth]{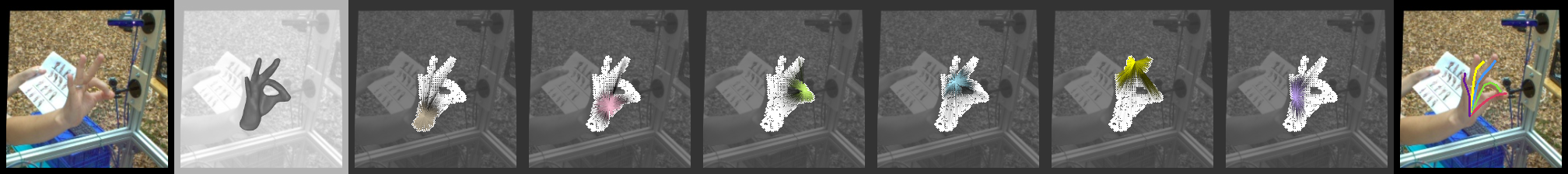}
    \includegraphics[width=1.0\linewidth]{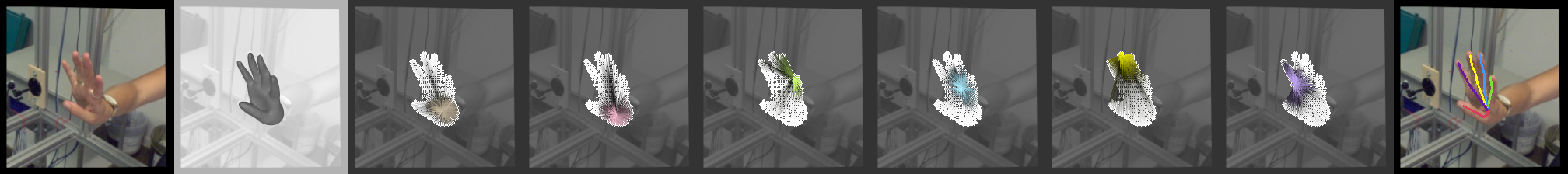}
    \includegraphics[width=1.0\linewidth]{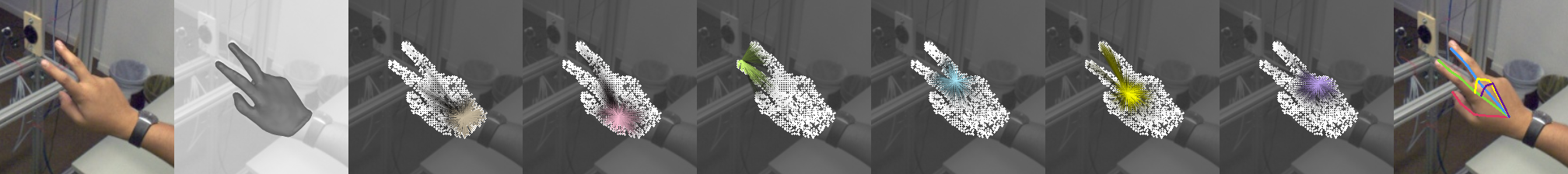}
    \includegraphics[width=1.0\linewidth]{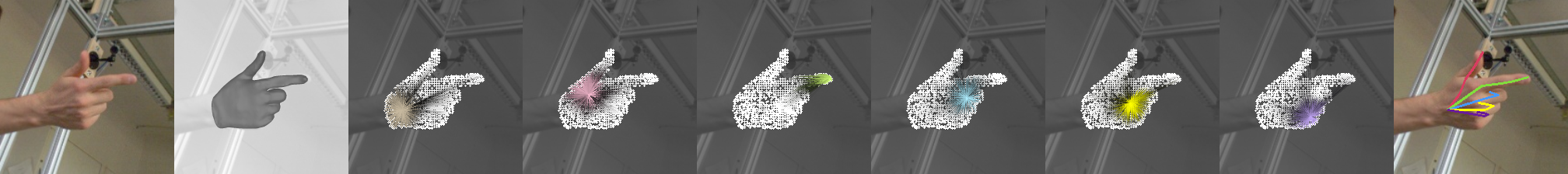}
    \includegraphics[width=1.0\linewidth]{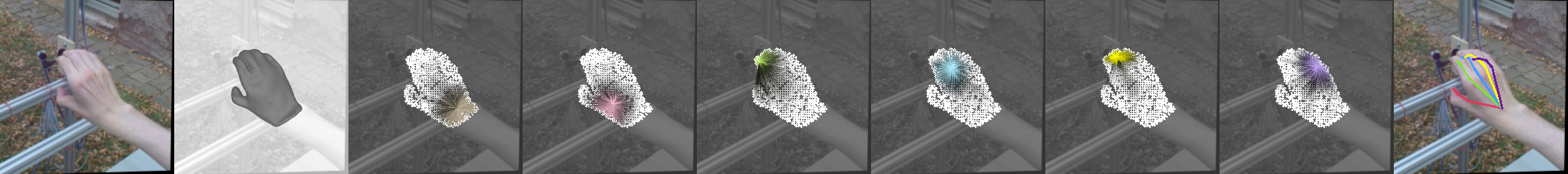}
    \includegraphics[width=1.0\linewidth]{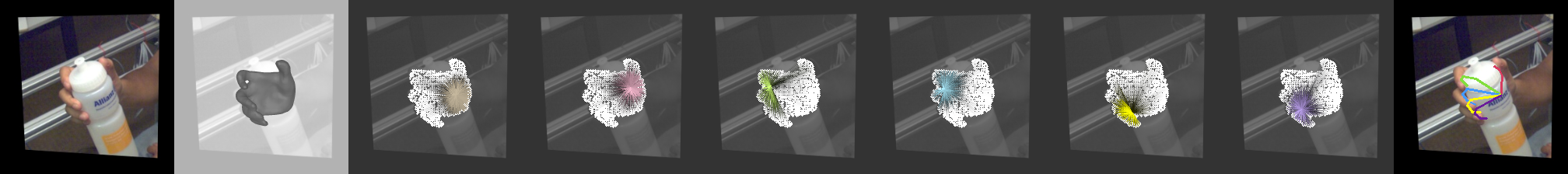}
    \vspace{-1ex}
    \includegraphics[width=1.0\linewidth]{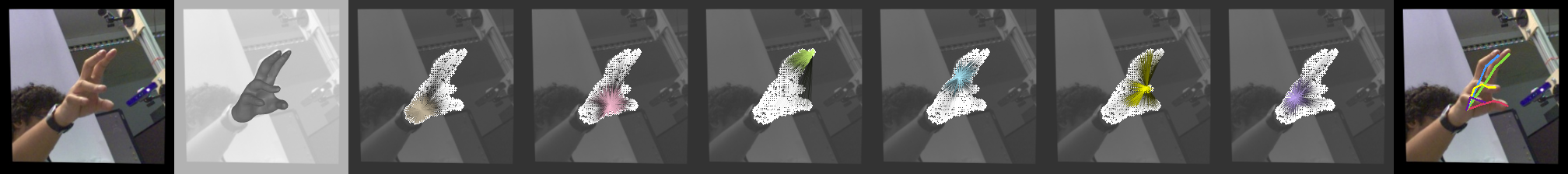}
        \hspace{-0.1cm} {\footnotesize (a)} \hspace{1.45cm}   {\footnotesize (b)}  \hspace{1.5cm}  {\footnotesize (c)} \hspace{1.4cm}  {\footnotesize (d)} \hspace{1.4cm}  {\footnotesize (e)} \hspace{1.5cm}  {\footnotesize (f)} \hspace{1.5cm}  {\footnotesize (g)} \hspace{1.3cm}  {\footnotesize (h)} \hspace{1.5cm}  {\footnotesize (i)}
    \vspace{-2ex}
    \caption{\label{fig:abs_fh_viz} \textbf{Additional qualitative results from CS-NVF for camera-space 3D hand pose on FreiHAND.} 
    CS-NVF can handle challenging cases of self-occlusion, occlusion by interacted object, and complex poses, leading to robust 3D hand pose. 
    Please refer to Sec.~\ref{sec:qualitative} for specific meaning of the rendering of camera-space 3D prediction results shown in columns from (b) to (i).
    }
	\end{center}
\end{figure*}

\begin{figure*}[t]
	\begin{center}
         \hspace{0.0cm} {\footnotesize RGB input} \hspace{0.3cm} {\footnotesize 3D hand mesh}  \hspace{0.7cm}  {\footnotesize Wrist} \hspace{0.7cm}   {\footnotesize Thumb root} \hspace{0.5cm}  {\footnotesize Index tip} \hspace{0.7cm}  {\footnotesize Middle root} \hspace{0.6cm}  {\footnotesize Ring tip} \hspace{0.8cm}  {\footnotesize Pinky root}  \hspace{0.35cm} {\footnotesize 3D hand pose}
    \includegraphics[width=1.0\linewidth]{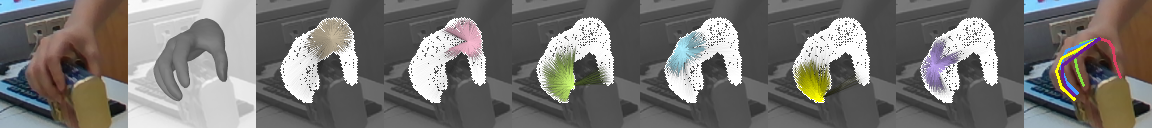}
    \includegraphics[width=1.0\linewidth]{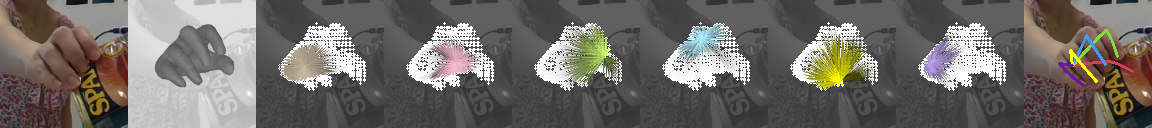}
    \includegraphics[width=1.0\linewidth]{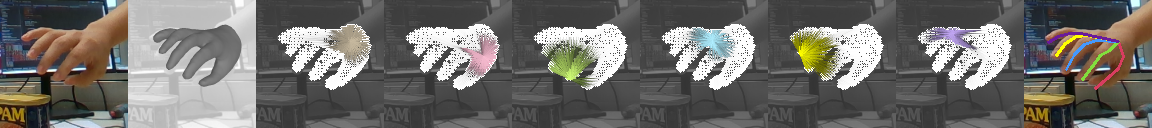}
    \includegraphics[width=1.0\linewidth]{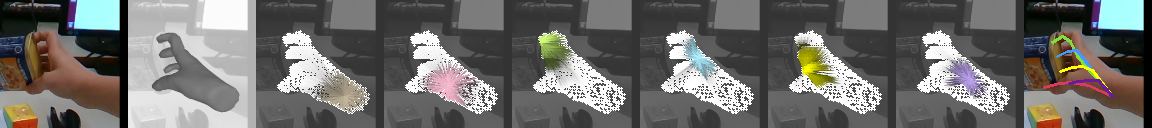}
    \includegraphics[width=1.0\linewidth]{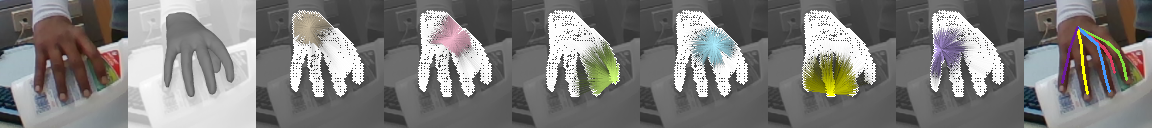}
    \includegraphics[width=1.0\linewidth]{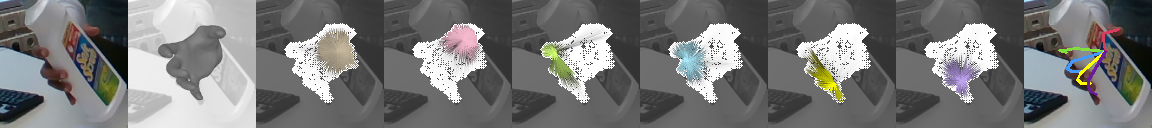}
    \includegraphics[width=1.0\linewidth]{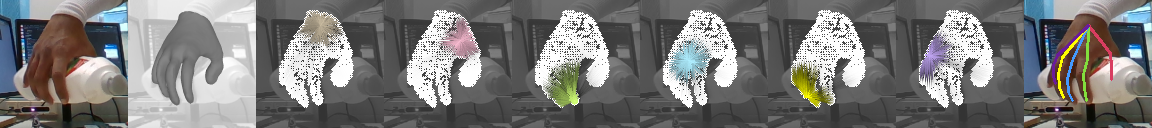}
    \includegraphics[width=1.0\linewidth]{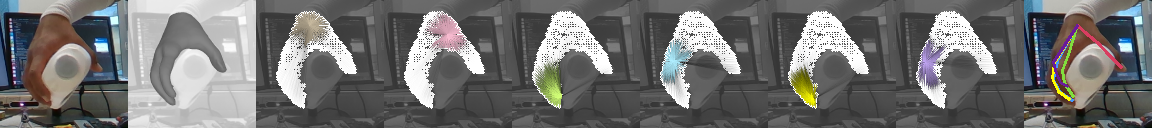}
    \includegraphics[width=1.0\linewidth]{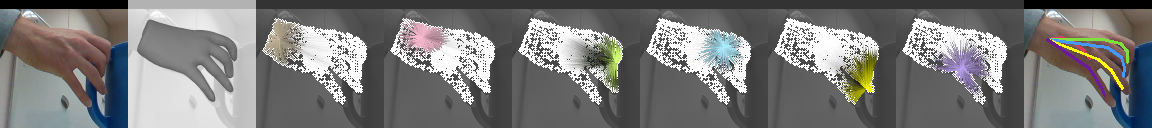}
    \includegraphics[width=1.0\linewidth]{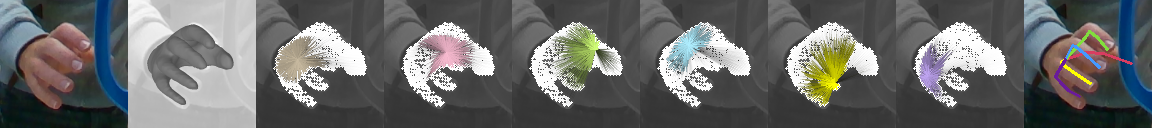}
    \vspace{-1ex}
    \includegraphics[width=1.0\linewidth]{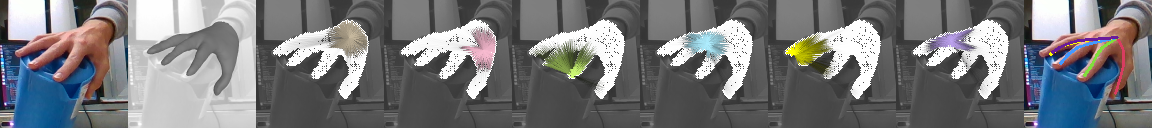}
        \hspace{-0.1cm} {\footnotesize (a)} \hspace{1.45cm}   {\footnotesize (b)}  \hspace{1.5cm}  {\footnotesize (c)} \hspace{1.4cm}  {\footnotesize (d)} \hspace{1.4cm}  {\footnotesize (e)} \hspace{1.5cm}  {\footnotesize (f)} \hspace{1.5cm}  {\footnotesize (g)} \hspace{1.3cm}  {\footnotesize (h)} \hspace{1.5cm}  {\footnotesize (i)}
    \vspace{-2ex}
	\caption{\label{fig:rel_fh_viz} \textbf{Additional qualitative results from RS-NVF for root-relative 3D hand pose on HO3D.}
    RS-NVF can handle challenging cases of self-occlusion and occlusion by interacted object, leading to robust 3D hand pose.
    Please refer to Sec.~\ref{sec:qualitative} for specific meaning of the rendering of root-relative 3D prediction results shown in columns from (b) to (i).
	}
	\end{center}
\end{figure*}

In Fig.~\ref{fig:abs_fh_viz} and Fig.~\ref{fig:rel_fh_viz}, we provide more qualitative results for CS-NVF for camera-space 3D hand pose estimation on FreiHAND and RS-NVF for root-relative 3D hand pose estimation on HO3D.

Given an RGB input, for each of the 3D query points densely sampled in camera frustum (CS-NVF) or hand root-relative 3D cube (RS-NVF), NVF regresses:
(i) the signed distance between the point and the hand surface; 
(ii) a set of 4D offset vectors. Each 4D offset vector consists of a 1D voting weight and a 3D unit directional vector from the query point to each hand joint, representing the closeness and direction from the point to each joint.
Following a vote-casting scheme, 4D offset vectors from near-surface points (\ie, points for which the predicted signed distance is below the clamping distance) are selected to calculate the 3D joint coordinates by a weighted average. Based on the overall pipeline, for each evaluation sample, we show:
\begin{itemize}
\item Column (a): single RGB input image.
\vspace{-1ex}
\item Column (b): 3D hand mesh generated by Marching Cubes~\cite{lorensen1987marching} from the signed distances predicted at 3D query points in the camera frustum (CS-NVF) or hand root-relative 3D cube (RS-NVF).
\vspace{-1ex}
\item Column (c-h): white circles as the valid 3D voters are the 3D points in the hand surface vicinity (\ie, points for which the predicted signed distance is below the clamping distance). Colored line denotes the predicted 1D voting weight from each near-surface 3D point (white circle) to a joint. Note that brighter line means larger weight, darker or no line means smaller weight.
\vspace{-1ex}
\item Column (i): estimated 3D pose via weighted average over predicted 4D offset vectors from near-surface points, following a vote-casting scheme.
\end{itemize}

\vspace{-2ex}
\customparagraph{Camera-Space 3D Hand Pose Estimation.}
Since CS-NVF generates all the predictions in the 3D camera space, all the results shown in column (b-i) of Fig.~\ref{fig:abs_fh_viz} are directly rendered by camera intrinsic parameters.

\customparagraph{Root-Relative 3D Hand Pose Estimation.}
For RS-NVF, since it generates all the predictions in the root-relative space, all the results in column (b-i) of Fig.~\ref{fig:rel_fh_viz} are first translated from the root-relative space into the camera space using the provided 3D ground-truth root location and then rendered by camera intrinsic parameters.

\customparagraph{Discussion.}
Based on qualitative results shown in Fig.~\ref{fig:abs_fh_viz} from CS-NVF and Fig.~\ref{fig:rel_fh_viz} from RS-NVF, we observe that:
\begin{itemize}
\vspace{-1ex}
\item Valid 3D voters: as indicated by the white circles in column (c-h) of both figures, among all the 3D points sampled at centers of voxels that fill up the camera frustum (CS-NVF) or hand root-relative 3D cube (RS-NVF) during inference, NVF is able to find points in the hand surface vicinity (\ie, points for which the predicted signed distance is below the clamping distance) even in occluded region and use these points as valid 3D voters for 3D pose estimation.
Different from classic pixel-level dense regression methods which mainly model foreground pixels, NVF shows its ability to reason points around the whole 3D hand surface.
\vspace{-1ex}
\item 1D voting weight predictions: as shown in column (c-h) of both figures, given the input images with  self-occlusion, occlusion caused by object, and complex poses, NVF is generally able to predict large voting weight for points close to hand joint and small value for points far away, resulting in well-shaped symmetric distribution around corresponding joint, even in occluded regions. Note that for each colored line, brighter line indicates larger voting weight, darker or no line indicates smaller voting weight. 
This demonstrates NVF's ability to model relation between each near-surface point and each hand joint.
\vspace{-1ex}
\item Signed distance predictions: as shown in column (b) of both figures for the hand mesh rendering (obtained using Marching Cubes from predicted signed distances), even in highly occluded region, NVF provides solid signed distance distribution showing its ability to reason the global hand structure/geometry.
\vspace{-1ex}
\item Estimated 3D hand pose: based on robust 3D point-wise predictions, NVF can then recover accurate 3D hand pose in challenging cases with severe occlusion and complex poses as shown in column (i) of both figures. Overall, these help to verify that, through direct dense modeling in 3D domain, NVF can model 3D dense local evidence and also the global hand structure/geometry, leading to robust 3D hand pose.
\end{itemize}

\vspace{-3ex}
\section{Qualitative Results on Real-World dataset}
\label{sec:gen}
\begin{figure*}[t]
	\begin{center}
    \includegraphics[width=1.0\linewidth]{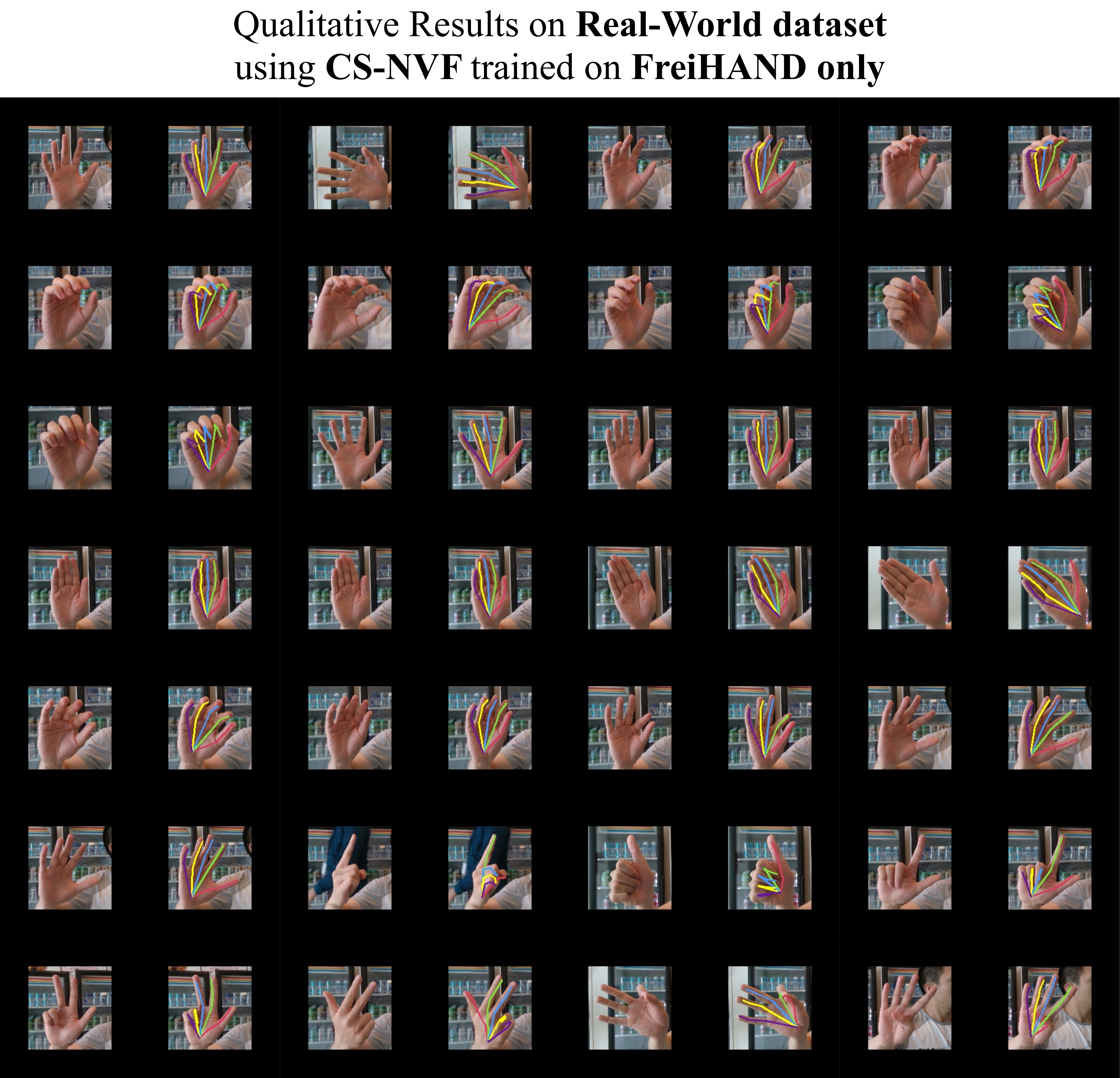}
	\end{center}
	\caption{\label{fig:gen} \textbf{Qualitative results on Real-World dataset~\cite{ge20193d} using CS-NVF trained on FreiHAND only.}
	For each pair of images, the left shows the input RGB image and the right shows the 3D hand pose directly rendered by camera intrinsic parameters. Our method shows its ability to generate solid results when testing on images in the wild/from another domain with various poses. This demonstrates, to some extent, the generalization capability of our proposed NVF.
	}
\end{figure*}

In Fig.~\ref{fig:gen}, we provide qualitative results on Real-World dataset~\cite{ge20193d} using our CS-NVF which is trained on FreiHAND dataset only. 
For each pair of images, the left shows the input RGB image and the right shows the 3D hand pose directly rendered by camera intrinsic parameters.
As shown in the figure, our method is able to generate solid results when testing on images in the wild/from another domain with various poses. This should demonstrate, to some extent, the generalization capability of our proposed NVF.

\customparagraph{Acknowledgments.}
This work is supported in part under the AI Research Institutes program by National Science Foundation and the Institute of Education Sciences, U.S. Department of Education through Award \# 2229873 - National AI Institute for Exceptional Education. Any opinions, findings and conclusions or recommendations expressed in this material are those of the author(s) and do not necessarily reflect the views of the National Science Foundation, the Institute of Education Sciences, or the U.S. Department of Education.

\end{document}